\documentclass{article}

\usepackage[preprint]{neurips_2026}

\usepackage[utf8]{inputenc} 
\usepackage[T1]{fontenc}    
\usepackage{hyperref}       
\usepackage{url}            
\usepackage{booktabs}       
\usepackage{amsfonts}       
\usepackage{nicefrac}       
\usepackage{microtype}      
\usepackage{xcolor}         
\usepackage{amsmath,amssymb,mathtools}
\usepackage{capt-of}
\usepackage{enumitem}
\usepackage[most]{tcolorbox}
\usepackage{tikz}
\usepackage{forest}
\usetikzlibrary{arrows.meta,positioning}
\usepackage{tikz}
\usetikzlibrary{positioning,fit,calc}
\usepackage[most]{tcolorbox}
\usepackage[ruled,vlined,linesnumbered,noend]{algorithm2e}
\usepackage{comment}
\usepackage{booktabs}
\usepackage{pgfplots}
\usepackage{subcaption}
\pgfplotsset{compat=1.18}
\usepackage{xcolor}
\definecolor{accblue}{HTML}{6C8EBF}
\definecolor{tokorange}{HTML}{F28E2B}
\definecolor{gridgray}{HTML}{EAEAEA}
\usepackage{graphicx}
\usepackage{wrapfig}
\usepackage{fancyvrb}
\usepackage{float}
\usepackage{tabularx}
\usepackage{caption}
\usepackage{amsmath, amssymb}
\usepackage{xcolor}
\usepackage[most]{tcolorbox}
\usepackage{enumitem}
\usepackage{listings}

\usepackage{float}
\definecolor{keywordcolor}{rgb}{0.7, 0.1, 0.1}   
\definecolor{commentcolor}{rgb}{0.4, 0.4, 0.4}   
\definecolor{stringcolor}{rgb}{0.5, 0.3, 0.2}    
\definecolor{symbolcolor}{rgb}{0.1, 0.2, 0.7}    
\definecolor{sortcolor}{rgb}{0.1, 0.5, 0.1}      
\definecolor{attributecolor}{rgb}{0.7, 0.1, 0.1} 
\definecolor{errorcolor}{rgb}{1, 0, 0}           

\definecolor{rawbg}{HTML}{F8FAFC}
\definecolor{rawframe}{HTML}{94A3B8}
\definecolor{procdbg}{HTML}{F0FDF4}
\definecolor{procframe}{HTML}{22C55E}
\definecolor{blockblue}{HTML}{2563EB}
\definecolor{blockpurple}{HTML}{7C3AED}
\definecolor{blockorange}{HTML}{EA580C}
\definecolor{blockgreen}{HTML}{16A34A}
\definecolor{blockteal}{HTML}{0F766E}
\tcbuselibrary{breakable}
\newtcolorbox{rawdatabox}[1][]{
  enhanced,
  breakable,
  colback=rawbg,
  colframe=rawframe,
  boxrule=0.6pt,
  arc=2mm,
  left=2mm,
  right=2mm,
  top=2mm,
  bottom=2mm,
  fonttitle=\bfseries,
  #1
}
\newtcolorbox{processedbox}[1][]{
  enhanced,
  breakable,
  colback=procdbg,
  colframe=procframe,
  boxrule=0.6pt,
  arc=2mm,
  left=2mm,
  right=2mm,
  top=2mm,
  bottom=2mm,
  fonttitle=\bfseries,
  #1
}
\newtcolorbox{semanticblock}[3][]{
  enhanced,
  breakable,
  colback=#2!4,
  colframe=#2,
  boxrule=0.6pt,
  arc=1.8mm,
  left=2mm,
  right=2mm,
  top=1.5mm,
  bottom=1.5mm,
  title={#3},
  fonttitle=\bfseries\small,
  #1
}

\newtcolorbox{promptbox}{
  breakable,
  colback=gray!8,
  colframe=black,
  boxrule=0.8pt,
  arc=2mm,
  left=6mm,
  right=6mm,
  top=2mm,
  bottom=2mm,
  width=\textwidth
}
\usepackage[most]{tcolorbox}
\tcbuselibrary{listings,breakable}
\usepackage{listings}
\lstdefinestyle{promptstyle}{
  language={},              
  mathescape=true,          
  basicstyle=\ttfamily\small,
  keywordstyle=\color{black},
  commentstyle=\color{black},
  stringstyle=\color{black},
  identifierstyle=\color{black},
  breaklines=true,
  breakatwhitespace=true,
  columns=fullflexible,
  keepspaces=true,
  showstringspaces=false,
  tabsize=2
}
\newtcblisting{promptbox1}{
  enhanced,
  breakable,
  colback=gray!5,
  colframe=black,
  boxrule=0.6pt,
  arc=2mm,
  left=2mm,
  right=2mm,
  top=1.5mm,
  bottom=1.5mm,
  listing only,
  listing options={style=promptstyle}
}
\newtcblisting{jsonbox}{
  enhanced,
  breakable,
  colback=gray!5,
  colframe=black,
  boxrule=0.6pt,
  arc=2mm,
  left=2mm,
  right=2mm,
  top=1.5mm,
  bottom=1.5mm,
  listing only,
  listing options={style=promptstyle,mathescape=true}
}
\title{CAM-Bench: A Benchmark for Computational and Applied Mathematics in Lean}

\author{
Wentao Long$^{1}$\thanks{Equal contribution.} \quad
Yunfei Zhang$^{2}$\footnotemark[1] \quad
Chenyi Li$^{3}$ \quad
Li Zhou$^{4}$ \quad
Chumin Sun$^{4}$ \quad
Zaiwen Wen$^{3}$ \thanks{Corresponding author: \texttt{wenzw@pku.edu.cn}.}
\\
$^1$Fudan University \quad
$^2$Qingdao University \quad
$^3$Peking University \quad
$^4$Huawei Technologies Ltd. \\
}

\begin{document}

\maketitle

\begin{abstract}
Formal theorem-proving benchmarks enable mechanically verifiable evaluation of mathematical reasoning in large language models. However, existing benchmarks mainly focus on Olympiad-style problems and algebraic domains, leaving computational and applied mathematics underrepresented. We introduce CAM-Bench \footnote{The CAM-Bench is open sourced at \url{https://github.com/optpku/CAM-Bench}.}, a Lean 4 theorem-proving benchmark of 1,000 Lean proof targets in computational and applied mathematics, with coverage spanning optimization, numerical linear algebra, and numerical analysis. These problems are adapted from textbook exercises and often depend on locally introduced definitions, notation, algorithms, and elementary results. To construct CAM-Bench, we develop a dependency-recovery pipeline that reconstructs the local textbook context needed to state each problem faithfully. It then normalizes each problem into a standalone informal theorem and translates it into a Lean target. We validate the resulting formal problems through Lean compilation and semantic review, checking both formal correctness and semantic alignment with the original exercises. For each problem, we release the raw exercise, recovered context, normalized informal theorem, and final Lean target. CAM-Bench complements existing formal mathematics benchmarks by targeting applied mathematics problems that rely on textbook concepts and elementary theorems, many of which are not directly available as standard Mathlib4 lemmas. We evaluate widely used large language models and formalization agents on CAM-Bench, and analyze common failure modes in tracking local assumptions, applying elementary results, decomposing proofs, and maintaining long-horizon control in Lean.
\end{abstract}

\section{Introduction}

Mathematical reasoning is a core capability for large language models (LLMs) and a central benchmark for reliable multi step inference.
Recent advances in reasoning models, proof search systems, and reinforcement learning with verifiable feedback have produced strong results on both informal mathematical benchmarks and formal theorem proving tasks \cite{Hend2021Math,he2024olympiadbench,guo2025deepseek,comanici2025gemini,chen2025seed,ren2025deepseek}.
Despite this progress, informal evaluation gives only a partial account of mathematical competence.
When the final answer is insufficient to validate the derivation, it is difficult to determine whether a model has actually followed a correct line of reasoning.
Formal proof assistants such as Lean \cite{moura2021lean}, Isabelle \cite{paulson1994isabelle}, Coq \cite{huet1997coq}, and HOL \cite{harrison1996hol} address this limitation by checking statements and proofs against a trusted kernel.
They therefore provide a rigorous way to evaluate whether model generated mathematics is not only plausible, but mechanically verifiable.

Existing formal reasoning benchmarks have been essential for measuring progress, yet they cover only part of the ability needed for formalization.
MiniF2F focuses on olympiad style statements across theorem proving systems \cite{zheng2021minif2f}.
ProofNet studies autoformalization for undergraduate mathematics \cite{azerbayev2023proofnet}.
PutnamBench targets competition level problems \cite{tsoukalas2024putnambench}.
FormalMATH broadens Lean~4 coverage through a large autoformalized corpus \cite{yu2025formalmath}.
More recent evaluations further show that success on curated formal benchmarks does not necessarily transfer to research level formalization projects or program verification tasks \cite{poiroux2025rlmeval,barkallah2025veribenchftp}.
These developments motivate benchmarks that test a different capability.
Existing benchmarks mainly focus on Olympiad style problems, algebraic domains, and isolated theorem statements, leaving computational and applied mathematics underrepresented.

This gap is important since many problems in computational and applied mathematics are not naturally stated as self contained theorem statements.
Textbook exercises in optimization, numerical linear algebra, and numerical analysis often depend on locally introduced definitions, notation, algorithms, chapter conventions, and elementary results.
These local dependencies may determine not only the meaning of the problem, but also the assumptions and proof structure needed for a faithful formal statement.
A benchmark in this setting therefore needs to evaluate whether models can track local assumptions, perform proof decomposition, and produce valid formal proofs.

We introduce CAM-Bench, a Lean~4 theorem-proving benchmark of 1,000 Lean proof targets in computational and applied mathematics.
The benchmark covers optimization, numerical linear algebra, and numerical analysis. Its problems are adapted from textbook exercises. For each problem, the local textbook context needed to state the exercise faithfully is recovered during dataset construction and exposed together with the Lean target.
Each task asks a model to construct a proof for a fixed Lean statement under its associated local context.
This design complements existing formal mathematics benchmarks by targeting applied mathematics problems that rely on textbook concepts and elementary theorems, many of which are not directly available as standard Mathlib4 lemmas.

To construct CAM-Bench, we develop a dependency recovery pipeline for context dependent textbook exercises. For each exercise, the pipeline reconstructs the surrounding mathematical context, including relevant definitions, notation, algorithms, assumptions, and elementary results. It then normalizes the exercise into a standalone informal theorem and translates it into a Lean~4 target. Compilation feedback and semantic review are used to check both the formal correctness of the Lean code and its semantic alignment with the original exercise. For each problem, we release the raw exercise, recovered context, normalized informal theorem, and final Lean target, together forming a transparent record of how the textbook exercise is converted into a proof task.

We evaluate widely used large language models and formalization agents on CAM-Bench. The resulting error analysis identifies failures in tracking local assumptions, applying elementary results, decomposing proofs, and maintaining long-horizon control in Lean, suggesting that context-dependent applied mathematics remains challenging even when the target statement is fixed and mechanically checkable. Our main contributions are as follows.
\begin{itemize}
    \item We introduce {CAM-Bench}, a Lean~4 theorem proving benchmark for computational and applied mathematics. The benchmark contains 1,000 Lean proof targets spanning optimization, numerical linear algebra, and numerical analysis, and provides traceable artifacts from the raw exercise to the final machine-checked Lean target.

    \item We design a dependency recovery and formalization pipeline for constructing faithful Lean targets from context-dependent textbook exercises. The pipeline recovers local mathematical context, decomposes heterogeneous exercises into semantic blocks, and uses compilation feedback, semantic review, and constrained repair to validate correctness and alignment.

    \item We evaluate representative LLM proving strategies on {CAM-Bench}, including direct proof generation, iterative repair, and agent-based proving. We further analyze their failures in formal representation, missing mathematical infrastructure, proof-artifact fidelity, and long-horizon proof control.
\end{itemize}

\begin{figure}[!t]
    \centering
    \includegraphics[width=1.0\textwidth]{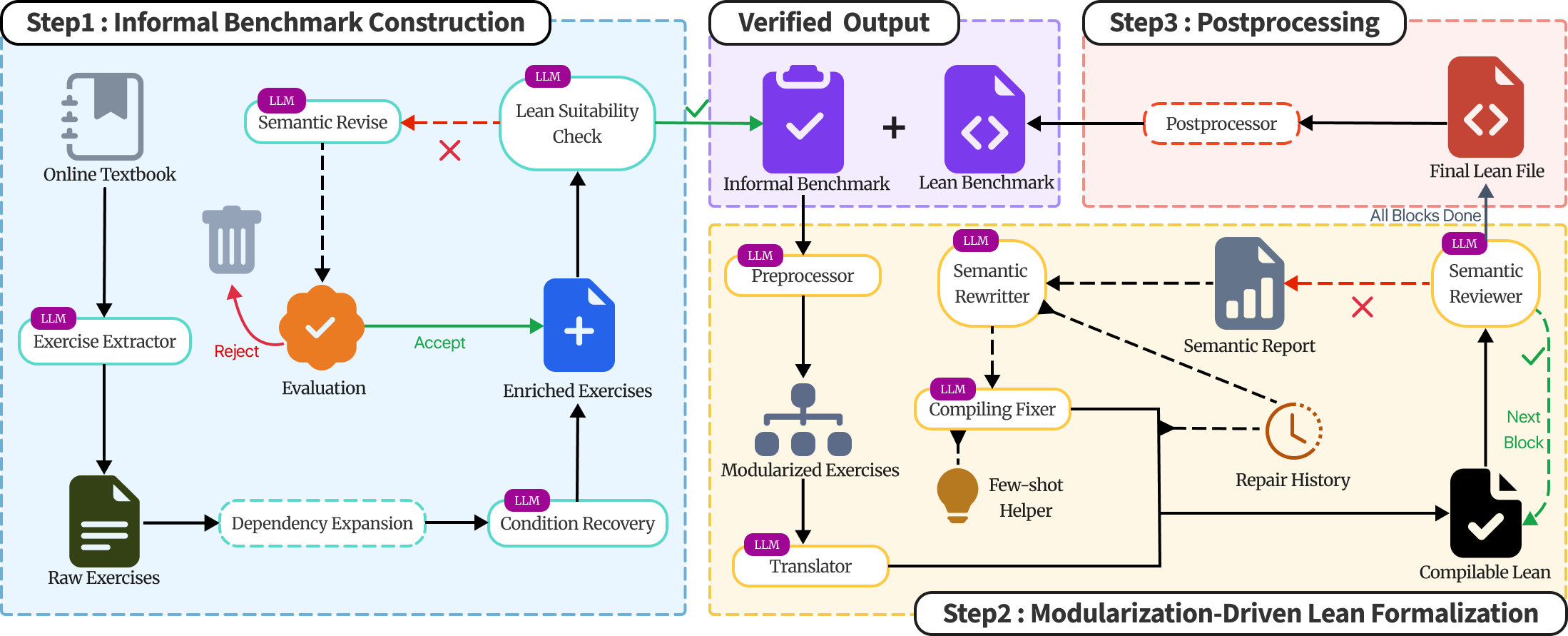}
    \caption{Pipeline Overview}
    \label{fig:Pipeline}
\end{figure}

\section{Preliminaries and Related Work}

\paragraph{Formal benchmarks.}
Formal benchmarks provide targets for measuring mathematical reasoning under proof assistant verification.
MiniF2F established a cross system testbed for olympiad style formal mathematics, and PutnamBench extends this line of evaluation to university level competition problems \cite{zheng2021minif2f,tsoukalas2024putnambench}.
ProofNet focuses on undergraduate mathematics by pairing informal problems with formal statements, making it an important resource for studying autoformalization \cite{azerbayev2023proofnet}.
Other benchmarks target specialized mathematical domains, including challenging informal to formal translation, combinatorial identities, and algebraic reasoning across multiple difficulty levels \cite{liu2023fimo,liu2025combibench,jiang2025fate}.
Large Lean corpora such as FormalMATH broaden coverage across mathematical areas and support more systematic evaluation of formal reasoning models \cite{yu2025formalmath}.
Recent evaluations also move beyond isolated theorem statements toward more realistic settings, including research level Lean blueprint projects and Lean~4 code verification tasks \cite{poiroux2025rlmeval,barkallah2025veribenchftp}.

\vspace{-0.6em}

\paragraph{Autoformalization for benchmark construction.}
Autoformalization translates informal mathematical language into formal statements and proofs, and large language models have made this direction increasingly practical \cite{wu2022autoformalization}.
Recent methods use retrieval, compiler feedback, process supervision, structured prompting, and iterative reflection to improve statement translation and proof generation \cite{weng2025autoformalization,lu2024process,lin2024lean,zhou2025solving}.
For benchmark construction, however, syntactic validity is only one requirement.
A Lean statement may compile while omitting an assumption, weakening the conclusion, changing a quantifier, or replacing the original exercise with a nearby theorem that is easier to prove.
Recent robustness studies further show that semantically similar paraphrases of the same informal statement can lead to different formalizations \cite{moore2025paraphrasing}.
\vspace{-0.3em}
\paragraph{Proof assistants and neural theorem proving.}
Systems such as Isabelle, Coq, HOL Light, and Lean have long supported large scale formal mathematics and program verification \cite{paulson1994isabelle,huet1997coq,harrison1996hol,de2015lean,moura2021lean}.
Recent neural theorem provers build on these environments by combining language model generation with search, retrieval, compiler feedback, and iterative repair.
Early systems showed that generative models can guide proof search in interactive theorem proving \cite{polu2020generative,lample2022hypertree}.
Subsequent Lean based provers make increasing use of retrieved premises, environment interaction, synthetic data, self correction, and reinforcement learning \cite{yang2023leandojo,xin2024deepseek,ren2025deepseek,wu2025internlm2,li2024hunyuanprover,xin2025bfs,lin2025goedel,wang2025kimina,chen2025seed,li2026optprover}.
Agent and project scale systems further emphasize long horizon coordination, tool use, and end to end formal development \cite{achim2025aristotle,wang2026m2f}. These systems have substantially improved formal proof generation. 

\section{Main Pipeline for Benchmark Construction}

\label{sec:main-pipeline}


The overall pipeline for constructing a paired informal--formal Lean benchmark is illustrated in Figure~\ref{fig:Pipeline}.
The system is organized into three major steps:
\textbf{(1) informal benchmark construction},
\textbf{(2) modularization-driven Lean formalization}, and
\textbf{(3) postprocessing}.
The first step converts raw textbook exercises into dependency-aware, self-contained informal statements.
The second step decomposes each verified statement into semantically typed blocks and incrementally translates them into Lean through compiler-guided and semantic repair.
The final step normalizes and packages the accepted Lean outputs into the final benchmark.

\subsection{Informal Benchmark Construction}
\label{subsec:informal-benchmark}

\paragraph{Exercise extraction.}
The pipeline begins by extracting raw exercise statements from online textbooks.
These raw exercises preserve the surface form of the original problems, but they are often not directly suitable for formalization.
They may refer to previous theorems, rely on chapter-level notation, or omit inherited assumptions.
Thus, the goal of this stage is not merely to extract exercises, but to transform them into self-contained formalization targets.
\vspace{-0.8em}
\paragraph{Recursive dependency expansion and enrichment.}
A central challenge in textbook-based benchmark construction is that exercise statements are rarely self-contained: a cited theorem may itself depend on earlier definitions, lemmas, equations, algorithms, or notation conventions, so resolving only explicitly mentioned references is insufficient for a Lean-suitable target.

To address this issue, we introduce a recursive dependency expansion and enrichment procedure, shown in Algorithm~\ref{alg:recursive-dependency-expansion}.
Starting from a raw exercise $x$, the procedure builds a bounded dependency tree $\mathcal{T}$ by repeatedly extracting references, retrieving the corresponding corpus items, and expanding their internal references up to depth budget $D$.
Each dependency node records its source text, reference label, hop depth, and provenance information, so each inserted item can be traced back to the dependency that contributed it.
Cyclic expansion is avoided by ignoring references already on the expansion path.

The tree is then processed bottom-up: each node merges the enriched representations of its children, and an LLM naturalizer rewrites the result to resolve local references, standardize notation, smooth transitions, and remove duplicated material, ensuring lower-level definitions are incorporated before they are used by higher-level dependencies.

At the root, a broader cleanup removes textbook-specific phrasing and pedagogical hints while preserving the mathematical objective.
An implicit-condition recovery step then makes required assumptions explicit---such as domain restrictions, positivity, dimensional compatibility, and convexity or regularity conditions---but only when supported by the original problem, recovered context, or well-formedness requirements of objects already present in the statement.
This enrichment is source-grounded rather than free-form: inserted definitions, equations, notation, and assumptions must be traceable to the source exercise or recovered dependencies; unresolved dependencies are held out rather than completed by conjecture.
The output $\widetilde{x}$ is a declarative, self-contained informal target that preserves the original assumptions, quantifiers, domains, and conclusion.
Additional details on statement cleaning, recursive dependency recovery, source-grounded enrichment, and implicit-condition completion are provided in Appendix~\ref{app:nl-intermediate-records}--\ref{app:self-contained-target-details}.

\begin{algorithm}[t]
\caption{Recursive Dependency Expansion and Enrichment}
\label{alg:recursive-dependency-expansion}
\KwIn{Exercise $x$, corpus $\mathcal{C}$, depth budget $D$}
\KwOut{Enriched exercise $\widetilde{x}$}

$\mathcal{T} \gets \textsc{Expand}(\textsc{NewNode}(x),\mathcal{C},D)$\;
$\widehat{x} \gets \textsc{Naturalize}(\mathcal{T}.root)$\;
\Return $\textsc{RecoverConditions}(\widehat{x},\mathcal{T})$\;

\SetKwFunction{Expand}{Expand}
\SetKwFunction{Naturalize}{Naturalize}
\SetKwProg{Fn}{Function}{:}{}

\Fn{\Expand{$v,\mathcal{C},d$}}{
    \If{$d=0$}{\Return $v$\;}
    \ForEach{$\rho \in \textsc{ExtractRefs}(\textsc{Text}(v))$}{
        $u \gets \textsc{RetrieveNode}(\rho,\mathcal{C})$\;
        \If{$u \neq \varnothing$ and $\rho \notin \textsc{Path}(v)$}{
            $\textsc{AddChild}(v,\textsc{Expand}(u,\mathcal{C},d-1))$\;
        }
    }
    \Return $v$\;
}

\Fn{\Naturalize{$v$}}{
    $\mathcal{M} \gets \{\textsc{Naturalize}(u):u\in\textsc{Children}(v)\}$\;
    $z \gets \textsc{Insert}(\textsc{Text}(v),\mathcal{M})$\;
    \Return $\textsc{LLM-Naturalize}(z,\textsc{IsRoot}(v))$\;
}
\end{algorithm}
\vspace{-1em}

\paragraph{Multi-round review and constrained repair.}
Since dependency enrichment is LLM-assisted, the resulting informal statements are passed through a validation loop before formalization.
Each enriched exercise is reviewed against the original source problem for missing assumptions, task drift, missing dependencies, unresolved definitions, and Lean suitability.
The instance is then classified as \textbf{accept}, \textbf{revise}, or \textbf{hold}.

Repair is used only for local, review-identified defects: it may normalize notation, inline already recovered dependencies, restore source-supported assumptions, or convert question-style prompts into theorem-style statements.
It may not introduce unsupported assumptions, invent missing dependencies, weaken or strengthen the claim, or change the proof objective.
Every repaired item is re-reviewed before proceeding to Lean formalization; cases with unrecovered dependencies are held out rather than completed by conjecture.
This loop prevents dependency enrichment from becoming unconstrained rewriting and keeps the informal benchmark faithful to the original textbook intent.
See Appendix~\ref{app:source-review-criteria} for the review criteria and decision labels, and Appendix~\ref{app:source-repair-rereview} for the constrained repair and re-review protocol.

\subsection{Modularization-Driven Lean Formalization}
\label{subsec:modularized-formalization}

\paragraph{Semantic block decomposition.}

\begin{wrapfigure}{r}{0.30\textwidth}
    \centering
    \vspace{-1.0em}
    \includegraphics[width=0.30\textwidth]{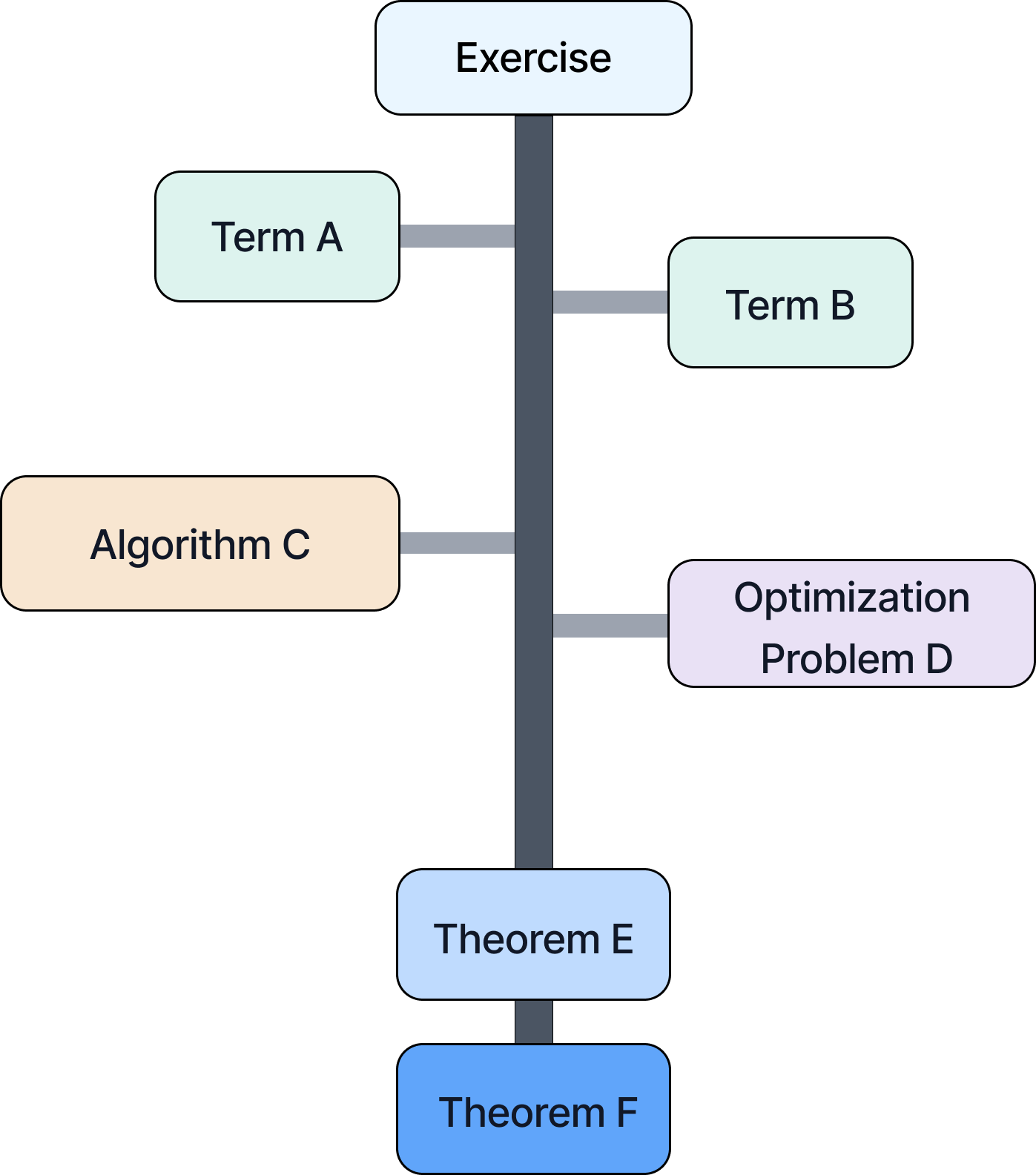}
    \vspace{-0.8em}
    \caption{\small Semantic block tree.}
    \label{fig:semantic-block-tree}
    \vspace{-2em}
\end{wrapfigure}

Textbook exercises are often heterogeneous: a single problem may combine definitions,
optimization objectives, algorithms, intermediate claims, hints,
and theorem targets.
Directly translating such mixed content into one Lean fragment can cause role
confusion and makes errors difficult to localize.

We therefore first convert each enriched exercise into a structured collection of
semantic blocks before Lean generation.
Each block is assigned a functional role according to how it should be used in the
formalization process.
For example, algorithmic descriptions and update rules are separated from
optimization formulations; formalizable concepts and assumptions are separated from
the final theorem target; and informal hints are retained as guidance but excluded
from the generated Lean declarations.
This decomposition turns a heterogeneous exercise into a theorem-centered dependency
structure. The target theorem serves as the root, while the remaining blocks serve as branches that provide 
supporting definitions, assumptions, constructions, or contextual information, as illustrated in Figure~\ref{fig:semantic-block-tree}.
By making the semantic role of each component explicit before Lean generation, later stages can focus on the relevant block rather than repeatedly rewriting the entire formalization.

\vspace{-0.5em}

\paragraph{Frozen-context block-wise construction.}
Lean formalization proceeds incrementally over the modularized blocks rather than by repeatedly regenerating the full Lean file.

At each step, the system focuses on a single target block, while previously verified declarations, assumptions, notation, and auxiliary definitions are kept fixed as read-only context.

This frozen-context protocol yields a stable, monotonically growing prefix: local edits cannot corrupt earlier blocks, cascading errors are reduced, and failures can usually be localized to the current block or its interface with the frozen prefix.

\vspace{-0.5em}

\paragraph{Translation, compilation repair, and semantic rewriting.}
For each target block, the translator generates Lean code according to the block type, using all previously accepted blocks as read-only context. The generated Lean code is then checked by the Lean compiler.

If the current block fails to compile, a compiler-guided fixer produces a local patch using the compiler diagnostics and the frozen context of previously accepted blocks.
Typical repairs include correcting syntax, adding missing type annotations, resolving namespace issues, or adjusting declarations to satisfy Lean's type system.
This compilation repair loop continues until the block becomes compilable or the repair budget is exhausted.

Every compilable candidate is then passed to a semantic reviewer.
The reviewer compares the Lean candidate with the corresponding block and checks whether the intended meaning is faithfully preserved.
If a mismatch is detected, the candidate is sent to a semantic rewriter, which produces a local semantic patch for the current block.
The patched code is then returned to the compilation repair loop to ensure that the semantic correction remains Lean-valid.
This combined loop ensures that accepted Lean blocks are both compilable and aligned with the informal statement.

\vspace{-0.7em}

\paragraph{History-aware repair and rejection control.}
Because iterative repair can oscillate, the system maintains a bounded history for each block, including recent attempts, compiler error signatures, failed patches, semantic review outcomes, and rejection reasons.
At each iteration, this history is provided together with the current feedback, making repair path-aware and discouraging previously failed edit directions.
In this way, failed attempts are not discarded, but are converted into negative constraints that guide subsequent repairs.

The pipeline also rejects rewrites that introduce semantic degradation, such as strengthening assumptions, weakening conclusions, dropping quantifiers, altering variable dependencies, changing notation in a meaning-changing way, or modifying optimization objectives.
This prevents Lean-valid patches from making the target easier by changing its mathematical content.
Together, history-aware repair and rejection control encourage monotonic progress toward blocks that are both compilable and faithful to the original textbook semantics.

\vspace{-0.7em}

\paragraph{Postprocessing.}
\label{subsec:postprocessing}

The final stage performs lightweight normalization and packaging.
Accepted Lean outputs are cleaned, organized, and assembled into final benchmark files for downstream evaluation. Additional details on semantic block decomposition, frozen-context construction, compiler-guided repair, semantic rewriting, history-aware rejection control, and postprocessing are provided in Appendix~\ref{app:semantic-block-decomposition}
--\ref{app:postprocessing-packaging}.

\section{Benchmark Analysis}

\begin{wrapfigure}{r}{0.50\textwidth}
    \centering
    \vspace{-1.5em}
    \includegraphics[width=0.45\textwidth]{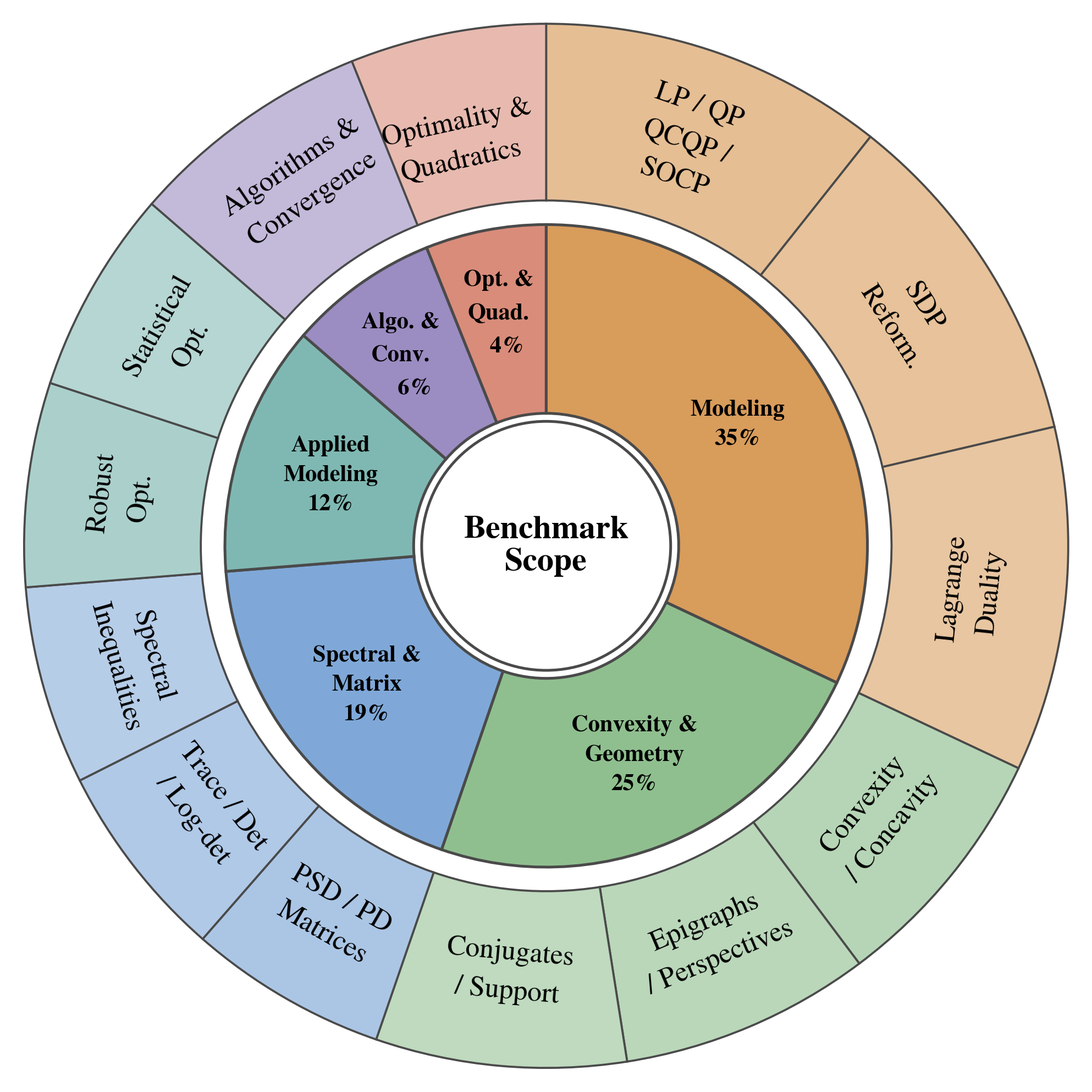}
    \vspace{0em}
    \caption{\small Distribution of benchmark targets}
    \label{fig:benchmark-taxonomy}
    \vspace{-1em}
\end{wrapfigure}

\subsection{Overall Dataset Statistics}

Our benchmark is a curated Lean 4 benchmark for
optimization-centric applied mathematics, comprising 1,000 Lean proof targets,
derived from 778 self-contained textbook exercises. Since a single exercise
may yield multiple formal theorem targets after decomposition, the number of
released Lean targets is larger than the number of source exercises; we refer to each of the 778 retained source exercises after dependency recovery and formalization as a \emph{benchmark instance}, so an instance may contain one or more proof targets. Each
retained target is processed through a hybrid construction pipeline involving
dependency completion, LLM-assisted semantic review, 
and Lean 4 formalization as discussed in Section \ref{sec:main-pipeline}.

The targets span six broad reasoning categories: spectral and matrix
reasoning; convexity and geometric inequalities; optimization modeling and
duality; optimality conditions and constrained quadratic reasoning; algorithms
and convergence; and applied modeling problems, including robust, statistical,
inverse, and network optimization as well as resource allocation. This
organization is based on the primary formal reasoning demand of each target
rather than on textbook chapter headings alone. Beyond topical breadth, the
benchmark is structurally challenging because many targets depend on
recovered local definitions, inherited assumptions, textbook-specific notation,
and auxiliary results that are not directly available as standard Mathlib4
lemmas.

\subsection{Preservation and Semantic Validation}
We assess construction quality at two levels. Source-level curation asks
whether the reviewed informal target remains faithful to the textbook exercise
and self-contained after dependency recovery and repair. Lean-level semantic
validation then asks whether the final Lean declaration preserves that reviewed
target after formal translation. The quantitative statistics below are derived from recorded semantic-review
logs produced by an independent LLM-as-a-judge reviewer and serve as diagnostic
signals for filtering and repair, rather than as the sole correctness criterion. At each semantic-review stage, every problem that enters the stage is assigned a structured review log recording the resulting status together with the supporting reasons; representative source-level and Lean-level logs are shown in Appendix~\ref{app:source-level-review-exp} and Appendix~\ref{app:example-semantic-repair}. We retain these logs as diagnostic construction records, and include representative
examples in the appendix to make the source-to-Lean decision process more auditable.

\begin{wrapfigure}{r}{0.45\textwidth}
    \centering
    \includegraphics[width=0.95\linewidth]{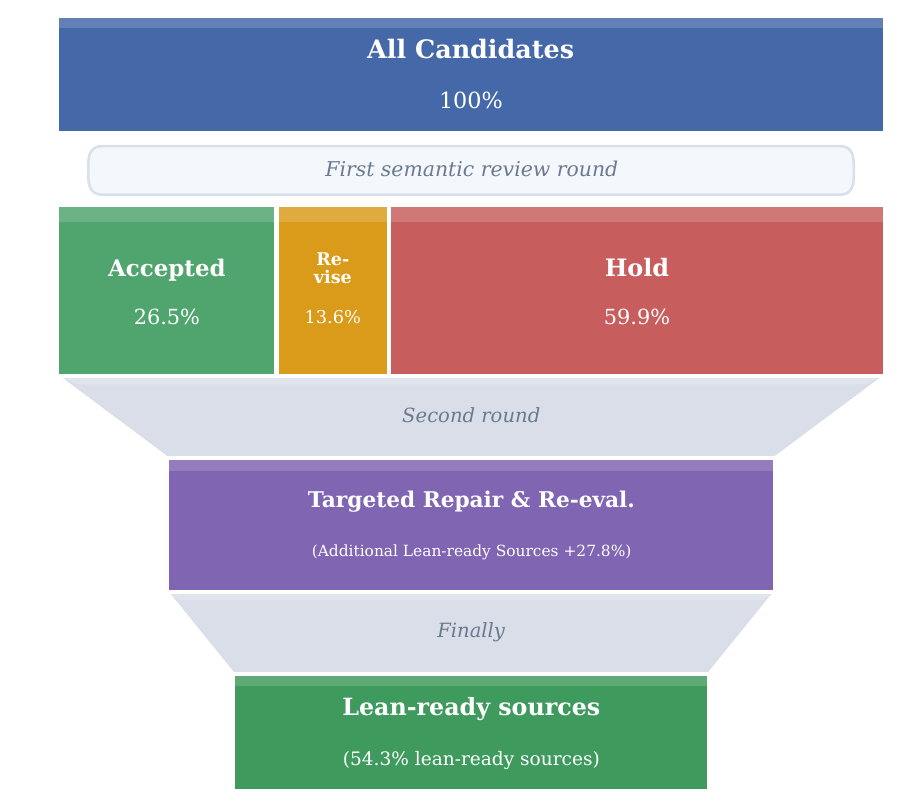}
    \vspace{0.2em}
    \caption{\small Source Review Funnel.}
    \label{fig:source-level_review}
    \vspace{-1.2em}
\end{wrapfigure}

\textbf{Source-level review.}
At the source level, we use the funnel-style process in Figure~\ref{fig:source-level_review} to  filter 
and recover candidate exercises.
In the first semantic review round, each candidate is assigned to one of three
outcomes: \emph{accept}, \emph{revise}, or \emph{hold}.
As shown in the figure, 26.5\% of candidates are accepted immediately, while
13.6\% are marked as \emph{revise} and 59.9\% as \emph{hold}.
Repair-eligible candidates from the revise branch and non-dependency hold
branch then undergo targeted repair and re-evaluation. This step recovers additional
Lean-ready informal targets, increasing the source-level usable yield from
26.5\% to 54.3\% before Lean formalization.

Candidates that remain unsatisfactory after this source-level review are excluded from the
Lean-ready source pool. This conservative funnel is designed to
preserve faithfulness to the original source exercise, self-containment,
dependency completeness, and suitability as Lean theorem targets.

\textbf{Lean-level validation.}
After formal translation, we perform a separate semantic validation stage for
the generated Lean declarations. Compilation with \texttt{sorry} checks
syntactic and type-theoretic well-formedness, but does not by itself ensure
semantic fidelity. We therefore compare each candidate Lean declaration
against its corresponding reviewed informal target for variables, assumptions,
quantifiers, objectives, constraints, and conclusions. Under the final
per-declaration review result, 92.5\% of candidate declarations are
classified as usable, while 7.5\% are classified as not usable and are
excluded from the released and evaluated benchmark unless corrected and re-validated. 
These percentages describe the candidate declaration pool at
this stage rather than the composition of the final released benchmark.

In this sense, the released benchmark is a conservatively retained subset of
the raw pipeline output rather than an automatic dump of all generated
formalizations. We report source-level validation details in  Appendix~\ref{app:source-level-review}
and Lean-level declaration validation details in Appendix~\ref{app:semantic-repair-loop}.

\section{Experiments}

\begin{figure}[ht]
    \centering
    \includegraphics[width=0.9\linewidth]{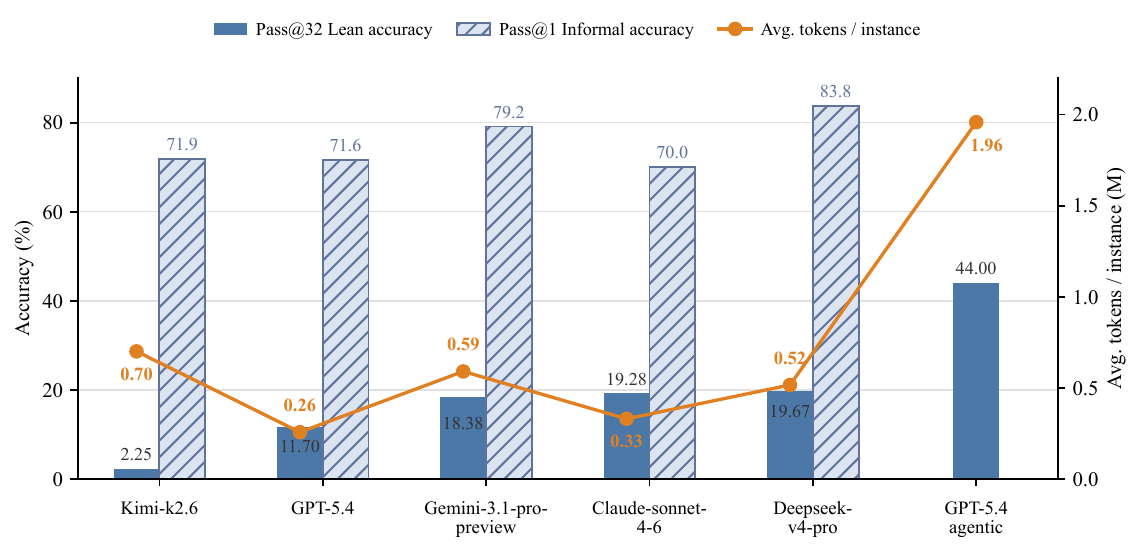}
    \caption{Model performance and token efficiency on CAM-Bench. Pass@32 Lean accuracy and Pass@1 informal accuracy are shown as percentages (\%) on the left y-axis, while average token usage per evaluated benchmark instance is shown in millions of tokens (M) on the right y-axis. Lean accuracy is evaluated at the file level, requiring all sorry blocks to be completed and the full Lean file to verify. The results highlight the gap between informal correctness and verified Lean success.}
    \label{fig:accuracy-token-comparison}
\end{figure}

We evaluate our benchmark using several frontier large language models. Figure~\ref{fig:accuracy-token-comparison} compares verified Lean accuracy, pass@1 informal accuracy, and average token consumption across frontier models and solving configurations.
All non-agentic baselines are evaluated under the same pass@32 protocol.
Success is measured at the file level: a benchmark instance is counted as solved only if all \texttt{sorry} blocks in the corresponding Lean file are correctly filled and the resulting file verifies ($N=778$ instances containing 1,000 \texttt{sorry} blocks in total).
We additionally include an agentic proof-search setting inspired by FormalProofBench~\citep{ravi2026formalproofbench}, where the model is given the informal problem, the fixed Lean statement, and the surrounding Lean context, and may iteratively use compiler feedback and Lean MCP tools before submitting a final proof.
Further details and interpretation of the experiment settings are given in Appendix~\ref{app:experimental-settings}.

\subsection{Benchmark Performance}
\label{subsec:benchmark-performance}
\paragraph{Overall performance.}
Among the non-agentic pass@32 baselines, \texttt{Deepseek-v4-pro}~\cite{deepseekai2026deepseekv4pro} achieves the highest verified Lean accuracy, reaching $19.67\%$.
It is closely followed by \texttt{Claude-sonnet-4.6}~\cite{anthropic2026claudesonnet46} at $19.28\%$.
\texttt{Gemini-3.1-pro-preview}~\cite{google2026gemini31propreview} obtains $18.38\%$, while \texttt{GPT-5.4}~\cite{openai2026gpt54} reaches $11.70\%$ and \texttt{Kimi-k2.6}~\cite{moonshotai2026kimik26} achieves $2.25\%$.
In contrast, the agentic repair harness substantially improves verified performance:
\texttt{GPT-5.4/agentic} reaches $44.00\%$ verified Lean accuracy, more than doubling the strongest non-agentic baseline.

\vspace{-0.5em}

\paragraph{Informal--formal gap.}
A clear gap appears between informal mathematical correctness and machine-checkable Lean correctness.
For the non-agentic baselines, pass@1 informal accuracy remains relatively high, ranging from $70.0\%$ to $83.8\%$.
However, their verified Lean accuracy remains much lower, with the strongest pass@32 baseline below $20\%$.
This discrepancy suggests that many model outputs capture the high-level mathematical argument but fail to satisfy the stricter requirements of Lean verification, including Lean syntax, Mathlib conventions, available hypotheses, and verifier constraints.

\vspace{-0.5em}

\paragraph{Agentic repair and token cost.}
The agentic configuration yields the largest gain in verified performance, improving \texttt{GPT-5.4} from $11.70\%$ under the pass@32 setting to $44.00\%$ under the agentic setting.
This improvement highlights the value of execution-aware refinement: compiler diagnostics, verifier failures, Lean MCP tool calls, and previous unsuccessful attempts provide useful signals for targeted proof repair.
The gain, however, comes at substantially higher cost: non-agentic baselines consume $0.26$--$0.70$M tokens per evaluated benchmark instance, whereas \texttt{GPT-5.4/agentic} requires about $1.96$M.
\subsection{Comparison on Agentic Proof Systems}
\label{subsec:agentic-proof-systems}

\begin{wrapfigure}{l}{0.45\textwidth}
    \centering
    \vspace{-0.8em}
    \includegraphics[width=0.95\linewidth]{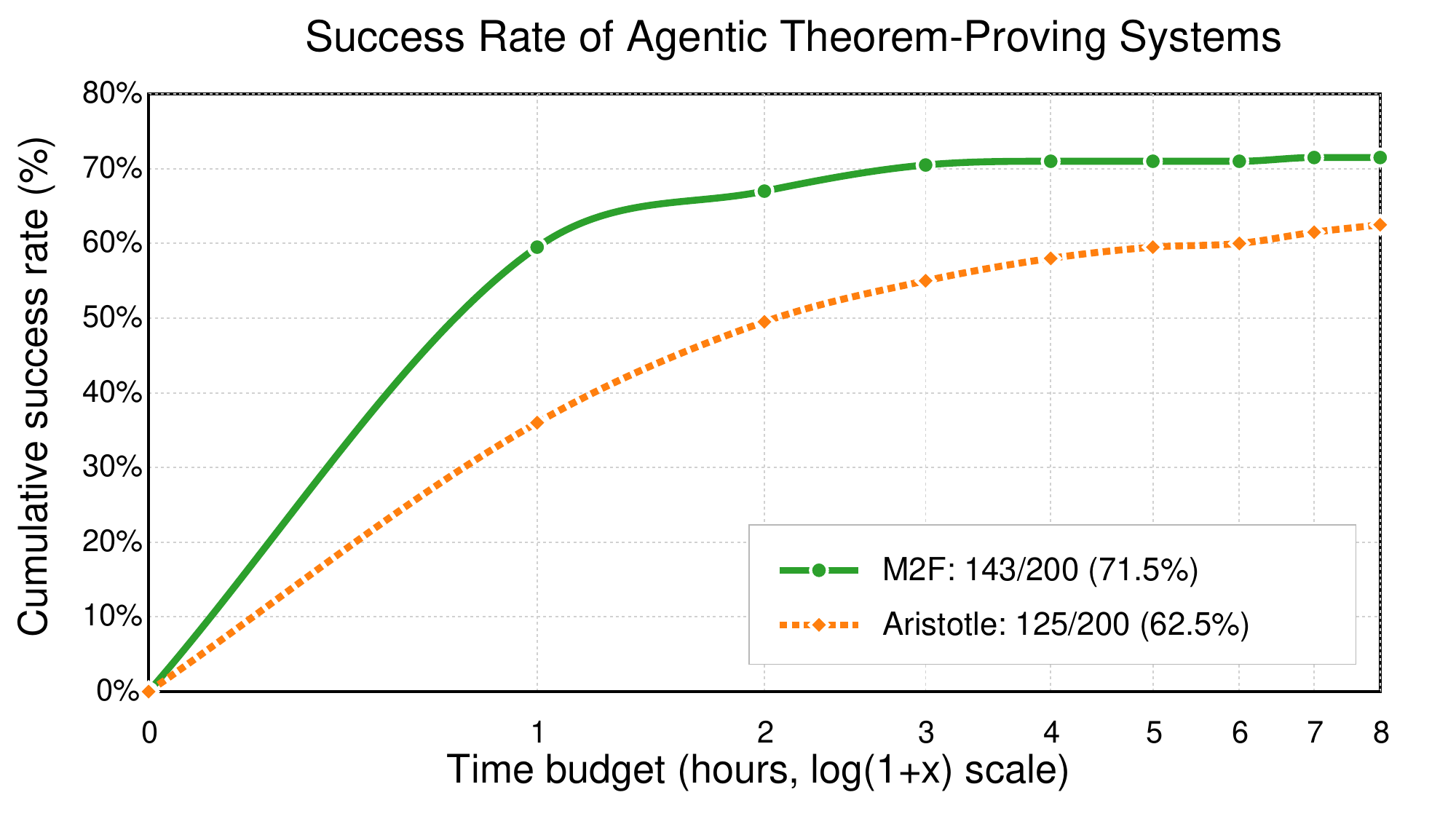}
    \vspace{-1.0em}
  \caption{\small Cumulative solved instances. The horizontal axis uses a pseudo-$\log(1+x)$ transform.}
    \label{fig:agentic-8h-success-curve}
    \vspace{-1.0em}
\end{wrapfigure}

We evaluate two external agentic theorem-proving systems, Aristotle
\cite{achim2025aristotle} and M2F \cite{wang2026m2f}, on 200 benchmark instances sampled uniformly at random from the released benchmark.
The same instance set, 8-hour wall-clock budget, and local Lean validation
script are used for both systems to ensure a fair comparison.
Figure~\ref{fig:agentic-8h-success-curve} shows the
cumulative number of instances that pass our strict validation protocol within the
8-hour budget; to make both the early-time behavior and the long tail visible,
the horizontal axis is displayed using a pseudo-$\log(1+x)$ transform.
Unlike pass@k decoding, these systems perform long-horizon proof
search with Lean compiler feedback and may construct auxiliary lemmas during
the solving process. We therefore adopt a strict single-problem external
validation protocol. An instance is counted as solved only if it is completed within
8 hours, the returned Lean project recompiles locally, preserves the original
theorem statements and existing definitions, and contains no pseudo-proof
artifacts such as \texttt{sorry}, \texttt{sorryAx}, \texttt{admit},
\texttt{axiom}, or unresolved \texttt{exact?}.

\paragraph{Agent performance.}
Under this protocol, M2F solves 143/200 sampled CAM-Bench instances
(71.5\%), while Aristotle solves 125/200 (62.5\%). Both substantially
outperform direct and lightweight-agentic baselines, showing the benefit of
long-horizon compiler-interactive proof search. However, the benchmark remains
unsaturated: 57 M2F instances and 75 Aristotle instances are still unsolved, often
due to missing mathematical infrastructure, long auxiliary-lemma chains, or
optimization-specific formal reasoning. Detailed validation counts are reported
in Appendix~\ref{app:m2f-aristotle-evaluation}.

\subsection{Error Analysis}
\label{subsec:error-analysis}

We analyze failed proof attempts from both the pass@32 LLM baselines and the
agentic proof-search systems, inspecting final Lean snapshots for pass@32 and final artifacts for the agentic setting. Since
a single failure may involve multiple symptoms, the labels are multi-label and
are used to summarize recurring bottlenecks rather than disjoint error
proportions. We group the dominant failure patterns into three categories.
Detailed definitions and fine-grained error statistics are provided in Appendix~\ref{app:error-analysis},
with representative examples in Appendix~\ref{app:formalization_failures}.
\paragraph{Library/API Grounding and Infrastructure.}
Errors in this category arise when the model fails to ground the proof in the
available Lean/Mathlib environment. They include hallucinated or incorrectly
used APIs, unavailable lemmas, missing bridge results, and failures to invoke
domain-specific infrastructure for matrices, spectra,
cones, convexity, optimality conditions, or duality.
\paragraph{Formal Representation and Type Discipline.}
Errors in this category are related to Lean's formal representation
requirements, including type mismatches, dimension or index errors, coercion
failures, missing typeclass assumptions, notation issues, and unresolved side
conditions.
\paragraph{Proof Decomposition and Long-Horizon Control.}
Errors in this category occur when the model fails to decompose a proof into
appropriate auxiliary lemmas or loses control over a long proof trajectory.
These failures often appear as unfinished helper-lemma chains, residual
placeholders, failed proof plans, or stalled reasoning in optimization, matrix,
convergence, and algorithmic arguments.

\begin{table}[ht]
\centering
\small
\caption{Coarse failure attribution for pass@32 LLM baselines and agentic proof-search systems. Labels are multi-label and therefore do not sum to 100\%.}
\label{tab:coarse-error-analysis}
\begin{tabular}{lccc}
\toprule
Failure attribution & pass@32 LLMs & M2F (N=57) & Aristotle (N=75) \\
\midrule
Library/API grounding and infrastructure
& 72.6\%
& 31/57 (54.4\%)
& 38/75 (50.7\%) \\

Formal representation and type discipline
& 50.8\%
& 26/57 (45.6\%)
& 29/75 (38.7\%) \\

Proof decomposition and long-horizon control
& 71.5\%
& 39/57 (68.4\%)
& 52/75 (69.3\%) \\
\bottomrule
\end{tabular}
\end{table}

Percentages are computed within failed proof blocks for pass@32 (pooled across all baseline models' failed attempts) and within  failed/repair cases for M2F ($N{=}57$) and Aristotle ($N{=}75$). Overall, the failures are rarely isolated syntax errors. As shown in Table~\ref{tab:coarse-error-analysis}, pass@32 baselines  often reach nontrivial proof attempts but get blocked by the
interaction of missing mathematical infrastructure, hallucinated or misused
Mathlib interfaces, and Lean type-system constraints, and insufficient proof decomposition. 
Agentic systems alleviate some local repair failures, but the remaining errors are still dominated by long-horizon proof construction and missing formal infrastructure. This suggests that CAM-Bench evaluates not only informal mathematical reasoning, but also the ability to ground applied-mathematics arguments in the actual Lean
environment and its available formal infrastructure.

\section{Conclusion}

We introduce CAM-Bench, a Lean 4 theorem-proving benchmark for computational and applied mathematics, constructed from textbook problems in optimization,
numerical linear algebra, and numerical analysis. To build it, we develop a
dependency-recovery and formalization pipeline that turns context-dependent exercises into self-contained Lean targets while preserving source traceability.
CAM-Bench remains challenging for current LLM-based theorem provers. Non-agentic
pass@32 baselines achieve limited verified accuracy, while agentic proof search
improves performance but leaves many targets unsolved. These results highlight
persistent bottlenecks in Mathlib grounding, Lean type discipline, and
long-horizon proof decomposition, and suggest that applied mathematics is a
valuable testbed for evaluating formal proof systems beyond isolated theorem
statements.

\bibliography{ref}
\bibliographystyle{plain}

\newpage

\appendix

\section{Details of Informal Benchmark Construction}
\label{app:nl-construction}

This appendix supplements Section~\ref{subsec:informal-benchmark} with
implementation details for the natural-language construction stage. While the
main text describes the overall extraction, recursive dependency expansion, and
review pipeline, this appendix specifies the intermediate records used to make
the process traceable and auditable.

The goal of this stage is to convert a context-dependent textbook exercise into
a self-contained informal mathematical target before Lean code is generated.
The input is a raw exercise statement extracted from an online textbook. The
output is \texttt{problem\_finally}, a Lean-facing informal target that preserves
the original mathematical objective while making assumptions, definitions,
domains, notation, and conclusions explicit.

\subsection{Intermediate Records}
\label{app:nl-intermediate-records}

Each exercise is stored as a sequence of intermediate JSON fields rather than as
a single rewritten statement. These fields record the main transformations from
the raw textbook exercise to the final informal target.

\begin{table}[h]
\centering
\small
\begin{tabular}{p{0.33\textwidth}p{0.57\textwidth}}
\toprule
Field & Role \\
\midrule
\texttt{problem} &
Raw textbook exercise and semantic anchor. \\

\texttt{problem\_clean} &
Cleaned mathematical task after removing non-essential hints, remarks,
background narrative, plotting instructions, or implementation-specific
guidance. \\

\texttt{problem\_with\_context} &
Cleaned task enriched with recovered textbook dependencies, such as referenced
definitions, equations, theorem statements, algorithms, notation conventions,
and surrounding assumptions. \\

\texttt{problem\_standardized\_math} &
Dependency-aware statement with normalized mathematical wording, consistent
notation, and clarified locally inherited objects. \\

\texttt{problem\_finally} &
Self-contained Lean-facing informal target used as input to later Lean
formalization. \\
\bottomrule
\end{tabular}
\caption{Intermediate natural-language fields used before Lean generation.}
\label{tab:nl-intermediate-fields}
\end{table}

These records make different sources of error easier to localize. Cleaning
errors can be identified by comparing \texttt{problem} and
\texttt{problem\_clean}. Dependency-recovery errors can be inspected through
\texttt{problem\_with\_context}. Normalization errors can be checked against
\texttt{problem\_standardized\_math}. Missing assumptions or unsupported
additions can then be reviewed in \texttt{problem\_finally}.

\subsection{Cleaning the Raw Exercise}
\label{app:raw-to-clean-details}

The first transformation produces \texttt{problem\_clean} from the raw extracted
exercise. This step is intentionally conservative. It removes non-essential
pedagogical material while preserving formulas, variables, constraints,
assumptions, conclusions, numbering, and explicit references.

The purpose of this step is not to solve, simplify, or reinterpret the problem.
Rather, it isolates the mathematical task from surrounding textbook exposition.
For example, hints, explanatory remarks, plotting instructions, and informal
implementation advice may be removed, but the mathematical objective and all
relevant hypotheses must remain unchanged.

Illustrative examples are provided in Appendix~\ref{app:raw_to_clean}, and the
corresponding cleaning prompt is shown in
Appendix~\ref{app:raw_to_clean_prompt}.

\subsection{Dependency recovery and Source-grounded enrichment}
\label{app:context-enrichment-details}

Starting from \texttt{problem\_clean}, the pipeline recovers local textbook
dependencies through the recursive dependency expansion procedure described in
Section~\ref{subsec:informal-benchmark}. The resulting field
\texttt{problem\_with\_context} records both the enriched exercise and the
dependency information used to construct it. This includes retrieved source
fragments, original reference labels, provenance information, and, when
applicable, multi-hop dependency expansions.

This enrichment step is source-grounded rather than free-form. The model may
incorporate retrieved definitions, equations, notation, or assumptions, but the
inserted information must be traceable to the source exercise or recovered
dependencies.

The enriched statement is then converted into
\texttt{problem\_standardized\_math}. This step folds the recovered context into
a more uniform mathematical statement. In particular, it may expand unresolved
references, standardize notation, clarify variables inherited from the textbook
environment, and remove duplicated inserted material. Its role is normalization
rather than mathematical rewriting: it must not introduce unsupported
assumptions, change the proof objective, or alter the original mathematical
content.

An illustrative example is provided in
Appendix~\ref{app:dep-rec-and-std-ex}.

\subsection{Constructing the Self-Contained Target}
\label{app:self-contained-target-details}

The field \texttt{problem\_finally} is the candidate self-contained informal
target passed to the next stage of the benchmark construction pipeline. It is
constructed primarily from \texttt{problem\_standardized\_math}, with
\texttt{problem\_with\_context} used only for explicitly recovered dependencies.

This step makes implicit conditions explicit when they are needed for a
well-formed Lean-facing statement. Examples include nonemptiness, nonzero
denominators, positivity, invertibility, dimensional compatibility, convexity or
regularity assumptions, and domain restrictions. Such conditions may be added
only when supported by the original problem, recovered context, or the
well-formedness requirements of objects already present in the exercise.

The output is a declarative informal target rather than a question-style
textbook prompt. However, this transformation is constrained: the final statement
must preserve the original assumptions, task objective, quantifiers, domains,
and conclusions.

An illustrative example is given in
Appendix~\ref{app:enriched_to_self-contained_exp}, and the prompt used for this
step is shown in
Appendix~\ref{app:enriched_to_self-contained_prompt}.

\subsection{Source-Level Review and Repair Interface}
\label{app:nl-review-interface}

Before Lean formalization, each \texttt{problem\_finally} is reviewed against
the original textbook problem. The review checks whether the candidate target
preserves the source assumptions, variables, domains, quantifiers, objectives,
constraints, and conclusions; whether all necessary definitions and dependencies
are present; and whether the statement is suitable as a Lean-facing theorem
target.

The reviewer assigns one of three labels: \texttt{accept}, \texttt{revise}, or
\texttt{hold}. Accepted items proceed to Lean formalization. Items marked
\texttt{revise} contain local repairable issues, such as a missing assumption,
a limited task drift, or a missing definition. These items enter a constrained
repair branch and are then submitted to re-review.

Items marked \texttt{hold} are handled separately because the label covers
different semantic risks. If the item depends on unresolved external context,
such as an unrecovered theorem, tagged equation, algorithm, or previous
exercise, we do not repair it by conjectural completion; the item is excluded
unless the missing context can be recovered from verified source material. If
the item is held only because of Lean-unsuitable wording or failed review
parsing, it may enter a separate hold-repair branch, provided that the original
source supports a conservative reformulation into a precise theorem-style
statement.

Repair is not an acceptance step. In both repair branches, only
\texttt{problem\_finally} may be modified, while earlier fields such as
\texttt{problem}, \texttt{problem\_clean}, and
\texttt{problem\_standardized\_math} remain fixed as audit anchors. The repair
may restore missing assumptions, make definitions explicit, inline already
recovered dependencies, or convert an open-ended exercise into an explicit
theorem-style claim. It may not introduce unsupported assumptions, weaken or
strengthen the mathematical claim, change the proof objective, or invent missing
dependencies. Every repaired item must pass re-review before it can be retained
for subsequent Lean formalization.

The full classification labels and  review protocol are reported in Appendix~\ref{app:source-level-review}.
The representative examples are given in Appendix~\ref{app:source-level-review-exp} and Appendix~\ref{app:repair_and_re-review_exp}, and the prompts used for these steps are shown in Appendix~\ref{app:source-level-review-prompt} and Appendix~\ref{app:repair_and_re-review_prompt}.
The statistics are shown in Appendix~\ref{app:source-review-statistics}.

\section{Source-Level Semantic Review}
\label{app:source-level-review}

This appendix describes the source-level semantic review procedure used during
benchmark construction. 

\subsection{Review Criteria and Decision Labels}
\label{app:source-review-criteria}

The source-level semantic reviewer compares \texttt{problem\_finally} against the
original \texttt{problem}. The goal is not to improve style, but to determine
whether the candidate target preserves the original mathematical meaning,
contains the assumptions and definitions needed for self-containedness, and is
suitable for Lean formalization. 

The review first records diagnostic issue flags. These flags identify the type
of semantic defect, if any, before the reviewer assigns a final decision label.
Table~\ref{tab:source-review-criteria} summarizes the issue categories used in
this stage.

\begin{table}[ht]
\centering
\small
\setlength{\tabcolsep}{5pt}
\renewcommand{\arraystretch}{1.12}
\caption{Source-level semantic review criteria and diagnostic issue flags.}
\label{tab:source-review-criteria}
\begin{tabularx}{\textwidth}{@{}p{0.20\textwidth}p{0.22\textwidth}X@{}}
\toprule
\textbf{Review aspect} & \textbf{Issue flag} & \textbf{Criterion} \\
\midrule
Assumption preservation
& \texttt{missing\_assumption}
& Flags cases where a necessary assumption from the source problem is absent or
changed in \texttt{problem\_finally}. This includes missing domain restrictions,
nonzero, nonnegative, positivity, positive-definiteness, index range,
continuity, differentiability, boundedness, invertibility, existence,
uniqueness, or other well-posedness assumptions. \\

Task preservation
& \texttt{task\_drift}
& Flags non-equivalent changes to the original mathematical task. This includes
changed quantifiers, changed logical direction, changed domains, weakened or
strengthened conclusions, dropped conclusions, added conclusions, or replacing
the source task by a nearby but different statement. Equivalent Lean-facing
reformulations are allowed only when they preserve the same domain,
quantifiers, and conclusion. \\

Self-containedness
& \texttt{missing\_dependency}
& Flags unresolved dependencies that prevent the statement from standing on its
own, such as missing equation tags, theorem labels, algorithms, previous
exercise references, local conventions, or cited facts that are mentioned but
not restated clearly enough. \\

Definition completeness
& \texttt{missing\_def}
& Flags technical terms, functions, operators, mappings, objects, or notation
that are used without definitions precise enough for formalization. This issue
is used only when the missing definition creates genuine ambiguity or blocks
formalization, not for harmless presentation-level refinements. \\

Formalization suitability
& \texttt{hold} condition
& Checks whether the task is suitable as a Lean-facing target. Proof-oriented
mathematical claims and precise computational statements are suitable.
Open-ended or presentation-oriented tasks such as ``describe'', ``formulate'',
``compute'', ``plot'', ``sketch'', or informal discussion are held unless they
have already been converted into a precise theorem-style statement. \\
\bottomrule
\end{tabularx}
\end{table}

\begin{table}[ht]
\centering
\small
\setlength{\tabcolsep}{6pt}
\renewcommand{\arraystretch}{1.12}
\caption{Decision labels for source-level semantic review.}
\label{tab:source-review-labels}
\begin{tabularx}{\textwidth}{@{}p{0.16\textwidth}X@{}}
\toprule
\textbf{Label} & \textbf{Assignment rule} \\
\midrule
\texttt{accept}
& Used only when no issue is present and the candidate target is suitable for
Lean formalization. The statement must preserve the original mathematical task,
be self-contained, and have a precise formalization-ready conclusion. \\

\texttt{revise}
& Used when the issue is local and repairable. In this workflow,
\texttt{revise} is triggered by \texttt{missing\_assumption},
\texttt{task\_drift}, or \texttt{missing\_def}, provided that the needed
dependency context is available and the task is otherwise formalizable. \\

\texttt{hold}
& Used when the candidate has unresolved dependencies or is not yet suitable for
Lean formalization. This includes missing external context, unresolved tags or
theorem references, open-ended exercise wording, non-formalizable tasks, or
failed/uncertain review outputs. \\
\bottomrule
\end{tabularx}

\end{table}

After these issue flags are recorded, the reviewer assigns one of three final
decision labels: \texttt{accept}, \texttt{revise}, or \texttt{hold}.
Table~\ref{tab:source-review-labels} gives the assignment rule for each
label.

The prompt for this stage is provided in
Appendix~\ref{app:source-level-review-prompt}. After review, each problem is accompanied by a review log that records its decision label and the reasons for that decision.
An illustrative example is
provided in Appendix~\ref{app:source-level-review-exp}.

\subsection{Constrained Repair and Re-review}
\label{app:source-repair-rereview}

Items marked \texttt{revise} are not rewritten freely. Instead, they enter a
constrained repair stage limited to the specific defect identified by the review
report. The repair step is not allowed to
introduce unsupported new assumptions, replace the original task by a nearby
easier statement, or alter the intended mathematical objective.

After repair, the updated \texttt{problem\_finally} is sent back to the same
source-level reviewer for re-evaluation. This second pass checks whether the
reported issue has been resolved without introducing a new one, and whether the
repaired target remains faithful, self-contained, and suitable for Lean
formalization.

Not every \texttt{hold} item enters the repair path. If the review report shows
that the candidate still depends on missing external context, such as an
unexpanded theorem reference, algorithm, or tagged formula, we do not attempt to
repair it by conjectural completion. Such cases are filtered out unless the
missing dependency can be recovered from verified source material. For
Lean-unsuitable cases, repair is considered only when the source exercise
already supports a conservative reformulation into a precise theorem-style
target; otherwise the item remains on hold and is excluded from the released
set.

The prompt for constrained repair and re-review is shown in
Appendix~\ref{app:repair_and_re-review_prompt}. An illustrative example is shown
in Appendix~\ref{app:repair_and_re-review_exp}.

\subsection{Global Source-Level Statistics}
\label{app:source-review-statistics}

We now summarize the aggregate outcomes of the source-level review, repair, and
re-review pipeline after data correction. The statistics are reported in three
steps, corresponding to three different input pools.

First, the original review is applied to the full source-level input pool. We
call this pool \texttt{OI}. Second, the first-round \texttt{revise} cases are
repaired and sent to re-review; this gives the \texttt{RI} pool. Third, the
first-round \texttt{hold} cases are filtered before re-review: cases held for
\texttt{missing\_dependency} are removed unless the missing context can be
recovered from verified source material. The remaining hold-derived pool is
called \texttt{HI}. Thus,
\[
\texttt{RI} = \texttt{R1 revise},
\qquad
\texttt{HI} = \texttt{R1 hold} - \texttt{missing\_dependency}.
\]

Step 1 reports the final decision distribution for each input pool. As shown in
Table~\ref{tab:source-review-round-outcomes}, the three pools have different
accept, revise, and hold rates, so we report them separately rather than merging
them into a single later-round result.

\begin{table}[H]
\centering
\small
\setlength{\tabcolsep}{9pt}
\renewcommand{\arraystretch}{1.12}
\caption{Outcome distributions for the three source-level review input pools
after data correction.}
\label{tab:source-review-round-outcomes}
\begin{tabular}{@{}lrrrr@{}}
\toprule
\textbf{Input pool} & \textbf{Input size} & \textbf{Accepted} & \textbf{Revised} & \textbf{Held} \\
\midrule
\texttt{OI} & 1550 & 410 (26.5\%) & 211 (13.6\%) & 929 (59.9\%) \\
\texttt{RI} & 211  & 93 (44.1\%)  & 67 (31.8\%)  & 51 (24.2\%) \\
\texttt{HI} & 578  & 338 (58.5\%) & 94 (16.3\%)  & 146 (25.3\%) \\
\bottomrule
\end{tabular}

\end{table}

Step 2 examines the cases marked \texttt{revise}. 
Table~\ref{tab:source-review-revise-reasons} shows the distribution of revise
reasons within each input pool. Since a single problem may carry more than one
issue label, the total number of issue labels can exceed the number of revised
cases.

\begin{table}[H]
\centering
\small
\setlength{\tabcolsep}{7pt}
\renewcommand{\arraystretch}{1.12}
\caption{Reason distributions within the \texttt{revise} subset of each
source-level review input pool.}
\label{tab:source-review-revise-reasons}
\begin{tabular}{@{}lrrrrr@{}}
\toprule
\textbf{Input pool}
& \textbf{Revised}
& \textbf{Missing assumption}
& \textbf{Task drift}
& \textbf{Missing definition} \\
\midrule
\texttt{OI} & 211 & 45 (21.3\%) & 118 (55.9\%) & 112 (53.1\%) \\
\texttt{RI} & 67  & 16 (23.9\%) & 47 (70.1\%)  & 29 (43.3\%) \\
\texttt{HI} & 94  & 16 (17.0\%) & 63 (67.0\%)  & 18 (19.1\%) \\
\bottomrule
\end{tabular}
\end{table}

Step 3 examines the cases marked \texttt{hold}. These cases are not treated as
ordinary local repair targets. Instead, they indicate unresolved dependencies,
Lean-unsuitable task forms, or parse failures. Table~\ref{tab:source-review-hold-reasons}
breaks down the hold reasons for each input pool.

\begin{table}[H]
\centering
\small
\setlength{\tabcolsep}{9pt}
\renewcommand{\arraystretch}{1.12}
\caption{Reason distributions within the \texttt{hold} subset of each
source-level review input pool.}
\label{tab:source-review-hold-reasons}
\begin{tabular}{@{}lrrrr@{}}
\toprule
\textbf{Input pool}
& \textbf{Held}
& \textbf{Missing dependency}
& \textbf{Lean-unsuitable}
& \textbf{Parse error} \\
\midrule
\texttt{OI} & 929 & 351 (37.8\%) & 512 (55.1\%) & 66 (7.1\%) \\
\texttt{RI} & 51  & 20 (39.2\%)  & 12 (23.5\%)  & 19 (37.3\%) \\
\texttt{HI} & 146 & 2 (1.4\%)    & 83 (56.8\%)  & 61 (41.8\%) \\
\bottomrule
\end{tabular}
\end{table}

\section{Details of Lean Formalization Pipeline}
\label{app:lean-formalization-pipeline}

This section supplements the Lean formalization pipeline in the main text.
We describe the record format, block-wise construction, semantic review, and
final packaging used to produce the benchmark artifacts.

\subsection{Lean-Facing Semantic Block Records}
\label{app:semantic-block-decomposition}

After preprocessing, each exercise is represented as an ordered list of
Lean-facing semantic records:
\[
\texttt{index},\quad
\texttt{source},\quad
\texttt{source\_idx},\quad
\texttt{kind},\quad
\texttt{content}.
\]
The optional \texttt{term} field is used when a block introduces a named
definition, optimization problem, or algorithm.
The field \texttt{kind} determines how the block is translated, as summarized in
Table~\ref{tab:semantic-block-types}.

\begin{table}[h]
\centering
\small
\begin{tabular}{p{0.14\linewidth}p{0.34\linewidth}p{0.38\linewidth}}
\toprule
\textbf{Kind} & \textbf{Semantic role} & \textbf{Lean-level treatment} \\
\midrule
\texttt{hints}
& Informal guidance.
& Kept for provenance, usually excluded from Lean generation. \\

\texttt{defn}
& Definitions, predicates, notation, or named terms.
& Translated into Lean definitions or auxiliary declarations. \\

\texttt{opt\_prob}
& Objectives, feasible sets, constraints, or optimality conditions.
& Translated into structures, feasibility predicates, or objective functions. \\

\texttt{algo} / \texttt{alg}
& Algorithms, updates, or iterative constructions.
& Translated into step functions, states, invariants, or validity predicates. \\

\texttt{thm}
& Theorem, lemma, proposition, or final target.
& Translated into a Lean theorem or lemma with proof body initialized as
\texttt{by sorry}. \\
\bottomrule
\end{tabular}
\caption{Lean-facing semantic block records.}
\label{tab:semantic-block-types}
\end{table}

A flattened theorem record has the following form:
\begin{verbatim}
{
  "index": 3,
  "source": "book/convex_optimization_Chapter3",
  "source_idx": "Exercise 3.24-(c)",
  "kind": "thm",
  "content": "Let a_1,...,a_n be real numbers with a_1 < ... < a_n.
              Prove that f is convex, concave, quasiconvex, and
              quasiconcave."
}
\end{verbatim}

Records are processed in \texttt{index} order.
The orchestrator inserts immutable block-boundary comments into the Lean file,
so each generated declaration can be traced back to its semantic record.

\subsection{Block-Wise Construction and Compilation Repair}
\label{app:blockwise-construction-repair}

Lean generation proceeds block by block.
For each target block, the prefix of already accepted Lean code is treated as a
frozen context; the model may only generate or repair the current block.
The combined file is checked by:
\begin{verbatim}
lake env lean --threads=<LEAN_THREADS> <file>
\end{verbatim}
For theorem-like blocks, \texttt{by sorry} is allowed during construction, so
compilation validates the statement and declaration interface rather than a
completed proof.

Algorithm~\ref{alg:blockwise-compilation-repair} summarizes the translation-repair
loop.
The few-shot helper retrieves similar translation or repair examples according
to the current block and, during repair, the Lean diagnostics.

A compact translation prompt is shown in Appendix~\ref{app:compact-translation-prompt}.

\begin{algorithm}[t]
\DontPrintSemicolon
\SetAlgoLined
\caption{Block-wise Lean translation with local repair}
\label{alg:blockwise-compilation-repair}
\KwIn{Semantic blocks $\mathcal{B}=(b_1,\ldots,b_m)$; Lean scaffold $L_0$;
repair budget $R$; few-shot helper $\mathcal{E}$}
\KwOut{Combined Lean file $L$}

$L \leftarrow L_0$\;

\For{$b \in \mathcal{B}$}{
  \If{$b$ is non-formalizable}{
    \textbf{continue}\;
  }

  insert immutable source comment for $b$ into $L$\;
  $F \leftarrow \textsc{FrozenContext}(L,b)$\;
  $E \leftarrow \textsc{RetrieveFewShot}(\mathcal{E},b)$\;
  $c \leftarrow \textsc{TranslateBlock}(b,F,E)$\;
  $L' \leftarrow \textsc{InsertBlock}(L,b,c)$\;

  \If{\textsc{Compile}$(L')$ succeeds}{
    $L \leftarrow L'$; freeze $b$; \textbf{continue}\;
  }

  \For{$r=1$ \KwTo $R$}{
    $D \leftarrow \textsc{Diagnostics}(L')$\;
    $E \leftarrow \textsc{RetrieveFewShot}(\mathcal{E},b,D)$\;
    $\widehat{c} \leftarrow \textsc{RepairBlock}(b,F,c,D,E)$\;
    $L' \leftarrow \textsc{ReplaceBlock}(L,b,\widehat{c})$\;

    \If{\textsc{Compile}$(L')$ succeeds}{
      $L \leftarrow L'$; freeze $b$; \textbf{break}\;
    }

    $c \leftarrow \widehat{c}$\;
  }

  \If{$b$ is not frozen}{
    mark $b$ as compilation failure and continue when possible\;
  }
}

\textsc{Compile}$(L)$\tcp*{final validation}
\Return $L$\;
\end{algorithm}

\subsection{Semantic Review and Alternating Repair}
\label{app:semantic-repair-loop}

Compilation does not ensure semantic faithfulness.
A compilable block may omit assumptions, alter quantifiers, change domains, or
weaken the intended conclusion.
Thus, each compilable block is reviewed against its semantic record.

The reviewer returns a structured report:
\begin{verbatim}
{
  "math_equivalent": true,
  "truth_judgement": "true|false|unknown|not_applicable",
  "overall_status": "usable|usable_with_revision|not_usable_yet",
  "dominant_issue_types": ["missing_assumption", "task_drift"],
  "top_priority_fix": "...",
  "issues": [
    {
      "severity": "P0|P1|P2",
      "issue_type": "missing_assumption|wrong_boundary_case|task_drift",
      "location": "current declaration",
      "reason": "...",
      "suggested_fix": "..."
    }
  ],
  "confidence": 0.0
}
\end{verbatim}

Blocks marked \texttt{usable} are accepted.
Otherwise, the current block is semantically rewritten, recompiled, and reviewed
again.
If semantic rewriting causes Lean errors, compilation repair is invoked under
the semantic constraint that the intended correction must be preserved.
Algorithm~\ref{alg:semantic-review-repair} summarizes this alternating loop.

\begin{algorithm}[t]
\DontPrintSemicolon
\SetAlgoLined
\caption{Block-wise semantic review and repair}
\label{alg:semantic-review-repair}
\KwIn{Lean file $L$; blocks $\mathcal{B}=(b_1,\ldots,b_m)$; budgets $S,R$;
few-shot helpers $\mathcal{H}$}
\KwOut{Reviewed Lean file $L$}

$\mathcal{A}\leftarrow[\,]$ \tcp*{accepted previous blocks}

\For{$i=1$ \KwTo $m$}{
  $b\leftarrow b_i$\;
  \If{$b$ is not formalizable}{\textbf{continue}\;}

  $F\leftarrow\textsc{FrozenContext}(L,b,\mathcal{A})$\;
  $\mathsf{Hist}_b\leftarrow[\,]$\;

  \For{$s=1$ \KwTo $S$}{
    $c\leftarrow\textsc{ExtractBlock}(L,b)$\;
    $\rho\leftarrow\textsc{Review}(b,c,F,\mathcal{A})$\;

    \If{$\rho$ accepts $c$}{
      append $b$ to $\mathcal{A}$; mark $b$ accepted; \textbf{break}\;
    }

    $c'\leftarrow\textsc{Rewrite}(b,c,\rho,F,\mathcal{A},\mathsf{Hist}_b)$\;
    $L'\leftarrow\textsc{PatchCurrentBlock}(L,b,c')$\;

    \If{\textsc{Compile}$(L')$ fails}{
      $\Delta\leftarrow\textsc{Diagnostics}(L')$\;
      $\mathsf{Hist}_b\leftarrow\mathsf{Hist}_b\cup\{(s,c',\rho,\Delta)\}$\;
      $\mathcal{E}\leftarrow\textsc{SelectFewShots}(\mathcal{H},\Delta,\rho)$\;
      $L'\leftarrow\textsc{CompilationRepair}
        (L',b,F,\mathcal{A},R,\Delta,\mathsf{Hist}_b,\mathcal{E})$\;
    }

    \eIf{\textsc{Compile}$(L')$ succeeds}{
      $L\leftarrow L'$ \tcp*{re-review repaired block next round}
    }{
      mark $b$ failed; \textbf{break}\;
    }
  }
}

\Return{$L$}\;
\end{algorithm}

\subsection{Postprocessing, Packaging, and Validation}
\label{app:postprocessing-packaging}

After semantic review, accepted Lean files are conservatively postprocessed and
exported.
This stage standardizes comments, namespaces, formatting, and metadata while
leaving accepted declarations unchanged.

Each benchmark instance contains the Lean file and a metadata record with the
source exercise, informal statement, block identifiers, validation status, and
basic construction statistics.
The pipeline also saves compiler diagnostics, semantic-review reports, repair
histories, and failure records.
The final artifacts separate statement construction from proof-generation
evaluation: downstream provers replace only the \texttt{sorry} placeholders
while the formalized statements remain fixed.

\section{Experimental Settings}
\label{app:experimental-settings}

\subsection{Code and Data Availability}
\label{app:code-data}

We provide the code and dataset to support reproducibility. The anonymized code repository is available at: \url{https://anonymous.4open.science/r/CAM-Bench-2D32}.

The dataset is hosted separately and can be accessed through the following DOI: \url{https://doi.org/10.7910/DVN/B5JTZI}.

\subsection{Baseline Setup}
\label{app:baseline-setup}

We evaluate Lean proof generation on CAM-Bench instances derived from textbook exercises in computational and applied mathematics. The benchmark contains 778 source exercises and 1,000 Lean proof targets,
each initialized with a \texttt{sorry} placeholder to be filled by the prover.

\paragraph{Pass@1 informal accuracy.}
Pass@1 informal accuracy is evaluated by prompting each model once per benchmark instance to produce a natural-language solution for the normalized informal theorem, without requiring Lean-valid syntax. An LLM evaluator then checks whether the returned argument correctly establishes the stated result; a response is counted as correct only if the core mathematical reasoning is valid and addresses the intended conclusion.

All Lean verification is performed with Lean 4 and Mathlib through Lake:
\[
\texttt{leanprover/lean4:v4.28.0}, \qquad
\texttt{mathlib:v4.28.0}.
\]
Each candidate proof is checked by running \texttt{lake env lean}. A proof attempt is considered successful only if the resulting Lean file compiles and passes integrity checks.

We compare the following baseline models:
\[
\texttt{gpt-5.4},\quad
\texttt{claude-sonnet-4-6},\quad
\texttt{kimi-k2.6},\quad
\texttt{deepseek-v4-pro},\quad
\texttt{gemini-3.1-pro-preview}.
\]
For the Lean proof-search experiments, we use temperature $0.4$, a maximum of 32 repair turns per \texttt{sorry} block, a file-level token budget of $10^6$, and a per-call output-token limit of 16,000.
Reported success is still computed at the file level: a benchmark instance is counted as solved only
when all \texttt{sorry} blocks in the corresponding Lean file are filled and the full file verifies. Lean MCP is enabled with Python transport, targeted tool mode, pool size 1, timeout 5 seconds, and the tools \texttt{lean\_unified\_search} and \texttt{lean\_leansearch}.

The proof-search solver processes each Lean file sequentially. It detects all \texttt{sorry} tokens, replaces the current \texttt{sorry} with a placeholder, asks the model to fill only that proof body, inserts the result, and verifies the prefix with Lean. Once a block is solved, the solver proceeds to the next \texttt{sorry}. Failed attempts are followed by retry prompts containing Lean error messages and, when available, MCP search context.

\subsection{Prompts for Pass@k Experiments}
\label{subsec:passk-prompts}

All pass@$k$ experiments use a fixed proof-generation prompt template.
The prompt provides the model with the informal problem, the target Lean
statement, and the surrounding Lean context, and asks it to return only a proof
body for the designated \texttt{sorry}.
The complete prompt templates and formatting constraints are given in
Appendix~\ref{app:passk-prompts}.

\subsection{Estimation for Pass@k Results}
\label{app:pass-k-estimation}

Table~\ref{tab:pass-k-results} reports the Pass@$k$ results for
$k \in \{1,2,4,8,16,32\}$.
These values summarize how performance changes as the sampling or repair budget
increases.  In particular, Pass@1 measures single-attempt success, while larger
values of $k$ measure the probability that at least one valid Lean solution is
found within a larger attempt budget.  Our main benchmark comparison uses
Pass@32, but the full Pass@$k$ curve provides a more detailed view of model
scaling under repeated attempts.

\begin{table}[t]
\centering
\small
\setlength{\tabcolsep}{8pt}
\renewcommand{\arraystretch}{1.12}
\caption{
Pass@$k$ results across different attempt budgets.
Pass@32 is the main setting used in our benchmark comparison.
}
\label{tab:pass-k-results}
\begin{tabular}{lcccccc}
\toprule
\textbf{Model}
& \textbf{Pass@1}
& \textbf{Pass@2}
& \textbf{Pass@4}
& \textbf{Pass@8}
& \textbf{Pass@16}
& \textbf{Pass@32} \\
\midrule
Claude-sonnet-4-6
& 5.14\% & 7.07\% & 9.51\% & 12.47\% & 15.42\% & 19.28\% \\
Deepseek-v4-pro
& 5.78\% & 8.48\% & 10.41\% & 12.85\% & 16.20\% & 19.67\% \\
Gemini-3.1-pro-preview
& 6.04\% & 8.87\% & 12.34\% & 14.27\% & 16.97\% & 18.38\% \\
GPT-5.4
& 2.96\% & 3.60\% & 5.53\% & 7.98\% & 10.55\% & 11.70\% \\
Kimi-k2.6
& 0.28\% & 0.28\% & 0.28\% & 0.56\% & 0.56\% & 2.25\% \\
\bottomrule
\end{tabular}
\end{table}

For each evaluation instance, we record whether the model solves all required
\texttt{sorry} blocks within at most $k$ attempts.  An instance is counted as
passed if every block is solved within the attempt budget and the final Lean code
passes compilation and integrity checks.  Thus, the entries in
Table~\ref{tab:pass-k-results} are empirical success rates under increasing
attempt budgets.

For a problem $i$, let $s_i(k)$ be the binary success indicator:
\[
s_i(k) =
\begin{cases}
1, & \text{if problem } i \text{ is solved within } k \text{ attempts},\\
0, & \text{otherwise}.
\end{cases}
\]
Given $N$ evaluated problems, the empirical Pass@$k$ is
\[
\mathrm{Pass@}k
=
\frac{1}{N}\sum_{i=1}^{N} s_i(k).
\]

When multiple independent samples are generated for the same problem, we use the
standard unbiased estimator.  If $n$ samples are generated and $c$ of them are
correct, then
\[
\widehat{\mathrm{Pass@}k}
=
1 -
\frac{\binom{n-c}{k}}{\binom{n}{k}},
\qquad n \ge k.
\]
This estimator gives the probability that at least one of $k$ sampled candidates
is correct.  In the sequential repair setting used by our Lean solver, the
reported Pass@32 corresponds to the empirical probability that the repair
trajectory succeeds within 32 turns.

\subsection{Agentic Proof-Search Setting}
\label{app:agentic-setting}

\begin{figure}[t]
    \centering
    \includegraphics[width=0.92\textwidth]{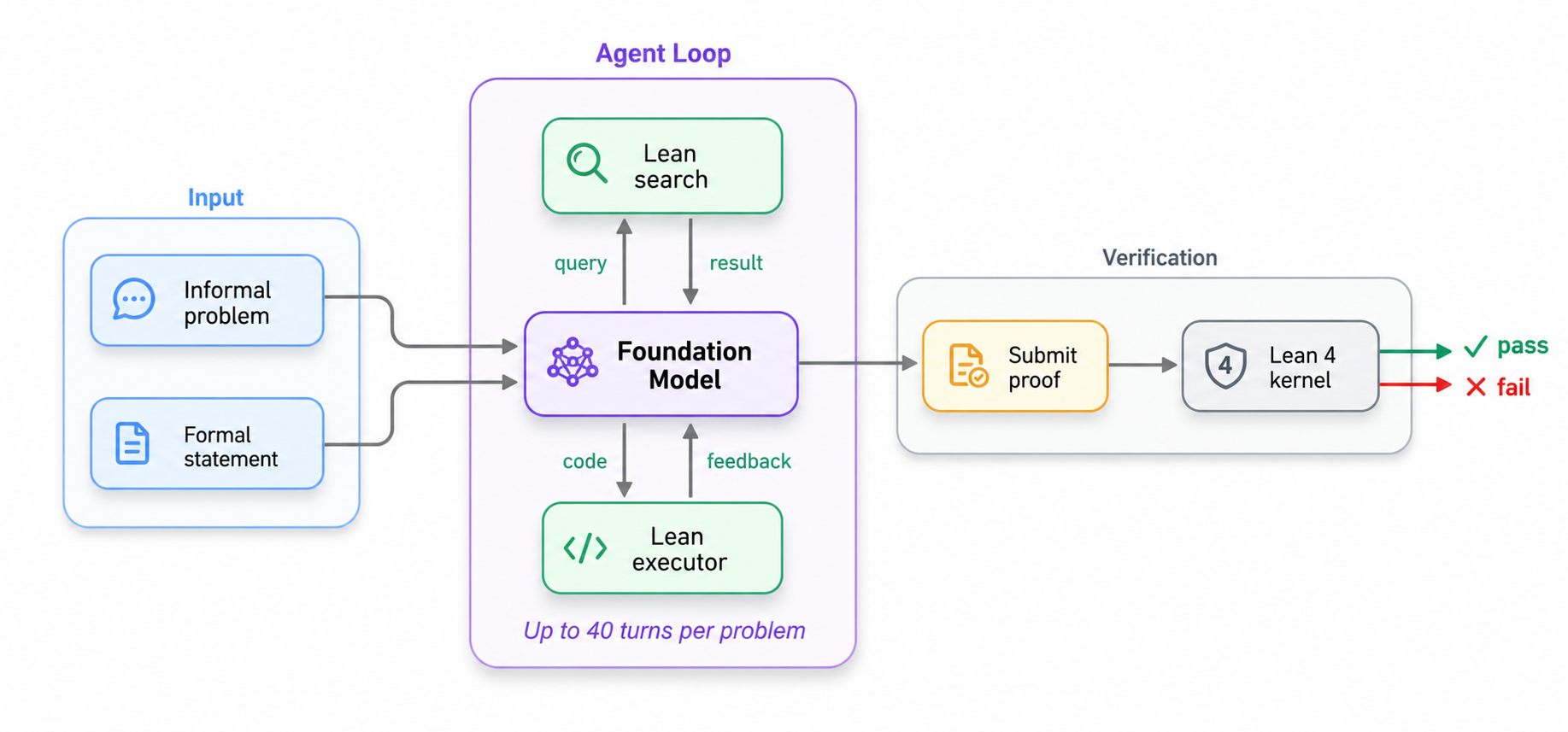}
    \caption{
    Agentic proof-search setting used in CAM-Bench.
    The model receives both the informal problem and the formal Lean target,
    iteratively interacts with Lean search and a Lean executor, and submits a proof
    for final Lean-kernel verification.
    Compared with the FormalProofBench~\citep{ravi2026formalproofbench} setup, our setting follows the same
    multiturn proof-search protocol but replaces Lean Loogle with Lean search.
    }
    \label{fig:agentic-proof-search}
\end{figure}

Figure~\ref{fig:agentic-proof-search} illustrates our agentic proof-search setting.
Our setting follows the multiturn, tool-augmented evaluation style of
FormalProofBench~\citep{ravi2026formalproofbench}, where the model is given the
natural-language problem, the Lean~4 formalization, and the required headers and
context, and may interact with proof-assistance tools before submitting a final proof.
While FormalProofBench uses Lean Loogle as its theorem-search tool, our
implementation uses \textbf{Lean search} instead.

For each problem, the agent receives the normalized informal theorem, the
corresponding Lean target, and all accepted local declarations recovered during
benchmark construction.
The agent may query Lean search for relevant Lean/Mathlib facts, test candidate
proof fragments with the Lean executor, and use compiler diagnostics, unsolved
goals, type mismatches, missing-lemma errors, and previous failed attempts to guide
iterative repair.
The agent is allowed up to 40 interaction turns per problem.
The final answer must be a proof body starting with \texttt{by}, which is inserted
into the original theorem without modifying the theorem statement or any preceding
definitions.

We use a strict verifier for final acceptance.
A task is counted as solved only if the completed Lean file recompiles successfully
and the submitted proof contains no prohibited proof shortcuts or unresolved holes,
including \texttt{sorry}, \texttt{admit}, \texttt{axiom}, \texttt{sorryAx},
\texttt{exact?}, or \texttt{apply?}.
We also require that the original theorem name, statement, assumptions, and
previously accepted declarations are preserved.
Thus, the agentic setting evaluates interactive proof construction rather than
statement rewriting or benchmark-specific weakening of the target.

\paragraph{Comparison with pass@32.}
The agentic results should be interpreted as a complementary evaluation setting rather than a direct replacement for the non-agentic pass@32 protocol.
In the pass@32 setting, each problem is evaluated through independent proof-generation attempts under a fixed sampling budget.
In contrast, the agentic setting allows the model to adaptively interact with Lean MCP tools, inspect compiler diagnostics, observe unsolved goals and type mismatches, and revise its proof attempts over multiple turns.
Thus, the two settings measure related but distinct capabilities: pass@32 primarily evaluates repeated direct proof generation, whereas the agentic setting evaluates interactive proof construction and execution-guided repair.

\paragraph{Performance--efficiency trade-off.}
The improvement of \texttt{GPT-5.4/agentic} suggests that many benchmark failures are repairable when the model receives structured feedback from the Lean environment.
Compiler errors can reveal local syntax issues, unavailable declarations, type mismatches, missing coercions, or mismatched hypotheses, while previous failed attempts help avoid repeated invalid proof patterns.
However, the agentic setting is not token-matched with the pass@32 baselines.
The agent spends additional tokens on tool calls, intermediate proof fragments, compiler outputs, and repair attempts.
In our experiments, pass@32 baselines use between $0.26$M and $0.70$M tokens per problem on average, whereas \texttt{GPT-5.4/agentic} uses about $1.96$M tokens per problem.
Therefore, we report agentic results separately and interpret them as evidence for the benefit of Lean-guided interactive repair, rather than as a strictly equal-budget comparison.

\section{Details of External Agent Evaluation and Failure Analysis}
\label{app:ext_agent_and_fail_anal}
\subsection{Details of M2F and Aristotle Evaluation}
\label{app:m2f-aristotle-evaluation}
This appendix reports the raw success counts for the M2F and Aristotle
evaluation. These counts correspond to the strict
externally validated results reported in the main text.
The 200 evaluation instances are sampled uniformly at random from the released benchmark and shared by both systems; the same instance  set, 8-hour wall-clock budget, and local Lean validation script are used for M2F and Aristotle to ensure a fair comparison. Because the evaluation is run on a 200 instances, the reported success rates should be interpreted as sample estimates for the full benchmark.
Table~\ref{tab:m2f-time-buckets} reports the M2F results, and
Table~\ref{tab:aristotle-time-buckets} reports the Aristotle results.

\begin{table}[ht]
\centering
\small
\caption{M2F results by solving-time bucket under the strict isolated 8-hour protocol.}
\label{tab:m2f-time-buckets}
\begin{tabular}{lrrrr}
\toprule
Time bucket & Total instances & Solved & Unsolved & Success rate \\
\midrule
$<30$ min & 107 & 102 & 5 & 95.3\% \\
30--60 min & 17 & 17 & 0 & 100.0\% \\
1--2 h & 18 & 15 & 3 & 83.3\% \\
2--4 h & 12 & 8 & 4 & 66.7\% \\
4--8 h & 4 & 1 & 3 & 25.0\% \\
8--12 h & 14 & 0 & 14 & 0.0\% \\
$>12$ h & 28 & 0 & 28 & 0.0\% \\
\midrule
Total & 200 & 143 & 57 & 71.5\% \\
\bottomrule
\end{tabular}
\end{table}

\begin{table}[ht]
\centering
\small
\caption{Aristotle results by solving-time bucket under the strict isolated 8-hour protocol.}
\label{tab:aristotle-time-buckets}
\begin{tabular}{lrrrr}
\toprule
Time bucket & Total instances & Solved & Unsolved & Success rate \\
\midrule
$<30$ min & 39 & 37 & 2 & 94.9\% \\
30--60 min & 42 & 35 & 7 & 83.3\% \\
1--2 h & 35 & 26 & 9 & 74.3\% \\
2--4 h & 32 & 18 & 14 & 56.2\% \\
4--8 h & 26 & 9 & 17 & 34.6\% \\
8--12 h & 15 & 0 & 15 & 0.0\% \\
$>12$ h & 11 & 0 & 11 & 0.0\% \\
\midrule
Total & 200 & 125 & 75 & 62.5\% \\
\bottomrule
\end{tabular}
\end{table}

\subsection{Details of Failure Analysis}
\label{app:error-analysis}
This appendix provides additional details for the formalization-error analysis
in Section~\ref{subsec:error-analysis}
. We analyze failed proof attempts from both
pass@32 LLM baselines and agentic proof-search systems. For pass@32, the unit
is a failed proof block and we inspect its final Lean snapshot. For agentic
systems, the unit is a non-clean run whose final artifact does not pass local
Lean compilation and validation. The labels are diagnostic rather than mutually
exclusive: a single failed attempt may involve multiple symptoms.

\paragraph{Library/API Grounding and Mathematical Infrastructure.}
This category covers failures where the model cannot ground a plausible
mathematical proof in the available Lean/Mathlib environment.

\begin{enumerate}
    \item \textbf{Hallucinated or invalid APIs.}
    The model invokes nonexistent theorem names, invalid fields, unsupported API
    patterns, or real declarations with incompatible signatures.

    \item \textbf{Missing bridge results.}
    The proof requires intermediate lemmas that connect the current goal to
    available Mathlib facts, but these bridge results are absent or not found.

    \item \textbf{Domain-specific infrastructure gaps.}
    The failure occurs in higher-level applied-mathematics infrastructure, such
    as matrices, spectra, positive semidefinite cones, convexity, optimality
    conditions, duality, or convergence.
\end{enumerate}

\paragraph{Formal Representation and Type Discipline.}
This category covers failures caused by Lean's precise representation
requirements.

\begin{enumerate}
    \item \textbf{Type, dimension, and index errors.}
    The generated proof uses incompatible types, dimensions, or indexing
    conventions, especially for vectors, matrices, finite index sets, and
    parameterized constraints.

    \item \textbf{Coercion and typeclass failures.}
    The proof misses required coercions, scalar casts, subtype conversions,
    order or normed-space instances, or other typeclass obligations.

    \item \textbf{Notation and side-condition failures.}
    The attempt uses unsuitable notation or leaves side conditions generated by
    rewriting, simplification, inequalities, or algebraic transformations
    unresolved.
\end{enumerate}

\paragraph{Proof Decomposition and Long-Horizon Control.}
This category covers failures where the proof requires a longer chain of
intermediate claims than the model can reliably organize.

\begin{enumerate}
    \item \textbf{Missing auxiliary lemmas.}
    The model does not introduce the intermediate claims needed to connect the
    assumptions to the final theorem.

    \item \textbf{Unfinished proof construction.}
    The final artifact still contains residual placeholders, incomplete helper
    lemmas, abandoned branches, or proof plans that exceed the search or repair
    budget.

    \item \textbf{Loss of long-horizon control.}
    The proof loses consistency across a multi-step argument involving
    optimization models, matrix identities, convexity reasoning, certificates,
    invariants, algorithms, or convergence claims.
\end{enumerate}

Overall, the observed failures are rarely isolated syntax errors. They usually
arise from the interaction of missing or misused Mathlib infrastructure, Lean
type-system constraints, and incomplete long-horizon proof planning.

We give three short case studies that illustrate the coarse error categories used in
Table~\ref{tab:coarse-error-analysis} in Appendix~\ref{app:formalization_failures}.

\section{Illustrative Examples}
\label{app:illustrative-example}

This appendix provides representative examples for the main construction stages
used in our benchmark pipeline. The examples are organized in the same order as
the pipeline: semantic block decomposition, raw-to-clean preprocessing,
dependency recovery and standardization, self-contained target construction,
source-level review, and constrained repair.

\subsection{Example: Semantic Block Decomposition}
\label{app:modularization-example}

This example illustrates how a heterogeneous textbook exercise is decomposed
into typed semantic blocks before Lean generation. The raw exercise combines
assumptions, notation conventions, an optimization formulation, a request to
derive KKT conditions, and a separate optimality claim. Directly translating
such mixed content into Lean is difficult because these components play
different formal roles and should not be treated as a single undifferentiated
problem statement.

Our preprocessing step separates the exercise into reusable mathematical
background, problem-specific data, and theorem targets. General concepts such
as KKT conditions, stationarity, primal feasibility, and complementary slackness
are extracted as definition blocks. The logarithmic simplex program is isolated
as an optimization-problem block, and the two requested tasks are separated into
theorem-target blocks.

\paragraph{Raw exercise.}
Let \(n\in\mathbf{N}\) with \(n\ge 2\). Let
\(a=(a_1,\dots,a_n)\in\mathbf{R}^n\) satisfy
\[
a_1 \ge a_2 \ge \cdots \ge a_n > 0,
\]
and define \(b=(b_1,\dots,b_n)\in\mathbf{R}^n\) by
\[
b_k=\frac{1}{a_k},\qquad k=1,\dots,n.
\]
Let \(\mathbf{1}\in\mathbf{R}^n\) denote the all-ones vector, and let
\(x\succeq 0\) mean \(x_i\ge 0\) for all \(i=1,\dots,n\).

The exercise considers the logarithmic simplex program
\[
\begin{array}{ll}
\text{minimize} &
-\log(a^T x)-\log(b^T x) \\[0.3em]
\text{subject to} &
x\succeq 0,\quad \mathbf{1}^T x=1,
\end{array}
\]
with variable \(x\in\mathbf{R}^n\). It then asks the solver to write the
Karush--Kuhn--Tucker conditions explicitly and to show that
\[
x=\left(\frac12,0,\ldots,0,\frac12\right)
\]
is optimal.

The KKT part of the exercise asks to show that the conditions are the existence
of a scalar multiplier \(\nu\in\mathbf{R}\) and inequality multipliers
\(\lambda=(\lambda_1,\dots,\lambda_n)\in\mathbf{R}^n\) such that
\[
\mathbf{1}^T x = 1,\qquad
x_i\ge 0,\qquad
\lambda_i\ge 0,\qquad
\lambda_i x_i=0
\quad (i=1,\dots,n),
\]
and, for each \(i=1,\dots,n\),
\[
-\frac{a_i}{a^T x}
-\frac{b_i}{b^T x}
+\nu-\lambda_i=0.
\]

\paragraph{Preprocessed semantic blocks.}
After preprocessing, the raw exercise is converted into the following typed
semantic blocks.

\begin{enumerate}[leftmargin=2.0em,itemsep=0.65em]

\item \textbf{Definition block: Karush--Kuhn--Tucker conditions.}
For a problem of minimizing \(f(x)\) subject to inequality constraints
\(g_i(x)\le 0\) and equality constraints \(h_j(x)=0\), the KKT conditions are
the existence of multipliers \(\lambda_i\ge 0\) and \(\nu_j\) such that primal
feasibility holds, stationarity holds,
\[
\nabla f(x)
+
\sum_i \lambda_i \nabla g_i(x)
+
\sum_j \nu_j \nabla h_j(x)
=
0,
\]
and complementary slackness holds:
\[
\lambda_i g_i(x)=0
\quad\text{for all } i.
\]

\item \textbf{Definition block: complementary slackness.}
For inequality constraints \(g_i(x)\le 0\) with multipliers
\(\lambda_i\ge 0\), complementary slackness means
\[
\lambda_i g_i(x)=0
\quad\text{for every } i.
\]

\item \textbf{Definition block: stationarity.}
A KKT point satisfies stationarity if the gradient of the Lagrangian with
respect to the primal variable vanishes:
\[
\nabla_x L(x,\lambda,\nu)=0.
\]

\item \textbf{Definition block: primal feasibility.}
A point \(x\) is primal feasible if it satisfies all original constraints of
the optimization problem.

\item \textbf{Optimization-problem block: logarithmic simplex program.}
The optimization problem is isolated as
\[
\begin{array}{ll}
\text{minimize} &
-\log(a^T x)-\log(b^T x) \\[0.3em]
\text{subject to} &
x\succeq 0,\quad \mathbf{1}^T x=1,
\end{array}
\]
with variable \(x\in\mathbf{R}^n\).

\item \textbf{Theorem-target block: explicit KKT conditions.}
Under the assumptions
\[
n\ge 2,\qquad
a_1\ge a_2\ge \cdots \ge a_n>0,\qquad
b_k=\frac1{a_k},
\]
the first theorem target asks for an explicit characterization of the KKT
conditions for the logarithmic simplex program. The target statement asserts the
existence of \(\nu\in\mathbf{R}\) and
\(\lambda=(\lambda_1,\dots,\lambda_n)\in\mathbf{R}^n\) satisfying
\[
\mathbf{1}^T x = 1,\qquad
x_i\ge 0,\qquad
\lambda_i\ge 0,\qquad
\lambda_i x_i=0
\quad (i=1,\dots,n),
\]
together with the stationarity equations
\[
-\frac{a_i}{a^T x}
-\frac{b_i}{b^T x}
+\nu-\lambda_i=0,
\qquad i=1,\dots,n.
\]

\item \textbf{Theorem-target block: optimality of the endpoint mixture.}
Under the same assumptions and for the same logarithmic simplex program, the
second theorem target asks to prove that
\[
x=\left(\frac12,0,\ldots,0,\frac12\right)
\]
is optimal.

\end{enumerate}

\paragraph{Discussion.}
This decomposition makes the role of each component explicit before Lean
generation. General mathematical background is separated from problem-specific
data, the optimization formulation is isolated from the requested claims, and
the two theorem targets can be translated, repaired, and validated
independently.

\subsection{Example: Raw-to-Clean Preprocessing}
\label{app:raw_to_clean}

This example illustrates the cleaning step that converts a raw textbook exercise
into the \texttt{problem\_clean} field. The cleaning step removes non-essential
pedagogical material while preserving the mathematical statement, notation,
assumptions, and objective.

\subsubsection{Removing Proof Hints}

\paragraph{Raw exercise.}
\begin{quote}
Exercise 11.5. When \(r:\mathbf{R}^n\to\mathbf{R}^n\), show that the function
\[
\phi(\lambda)
=
\left\|
  (J^T J+\lambda I)^{-1}J^T r
\right\|
\]
is monotonically decreasing in \(\lambda\), unless \(J^T r=0\).
Hint: Use the singular-value decomposition of \(J\).
\end{quote}

\paragraph{Cleaned exercise.}
\begin{quote}
Exercise 11.5. When \(r:\mathbf{R}^n\to\mathbf{R}^n\), show that the function
\[
\phi(\lambda)
=
\left\|
  (J^T J+\lambda I)^{-1}J^T r
\right\|
\]
is monotonically decreasing in \(\lambda\), unless \(J^T r=0\).
\end{quote}

\paragraph{Discussion.}
This example illustrates a minimal cleaning operation. The mathematical
statement, notation, and exception condition are preserved exactly, while the
proof hint is removed.

\subsubsection{Removing Application Narrative}

\paragraph{Raw exercise.}
\begin{quote}
Exercise 17.3. Flux balance analysis in systems biology. Flux balance analysis
is based on a simple model of reactions in a cell, keeping track only of the
gross rate of consumption and production of chemical species. Based on known
stoichiometry and upper bounds on some reaction rates, we can compute bounds on
other reaction rates or on cell growth.

We focus on \(m\) metabolites \(M_1,\ldots,M_m\) and \(n\) reactions
\(R_1,\ldots,R_n\), with nonnegative reaction rates \(v_1,\ldots,v_n\). The
stoichiometry matrix \(S\in\mathbf{R}^{m\times n}\) is defined so that
\(S_{ij}\) is the rate of production of metabolite \(M_i\) due to unit reaction
rate \(v_j=1\), with consumption represented by negative production.

The last reaction \(R_n\) corresponds to biomass creation, so \(v_n\) is the
cell growth rate. Conservation of metabolites is expressed as
\[
Sv=0.
\]
We are also given upper limits on some reaction rates, written as
\[
v\succeq 0,\qquad v\preceq v^{\max},
\]
where \(v_j^{\max}=\infty\) if no upper bound on reaction rate \(j\) is known.

The questions below pertain to the data found in \texttt{fba\_data.m}.

(a) Find the maximum possible cell growth rate \(G^\star\), as well as optimal
Lagrange multipliers for the reaction rate limits. How sensitive is the maximum
growth rate to the various reaction rate limits?
\end{quote}

\paragraph{Cleaned exercise.}
\begin{quote}
Exercise 17.3. Let \(S\in\mathbf{R}^{m\times n}\) be the stoichiometry matrix,
let \(v_1,\ldots,v_n\) be nonnegative reaction rates, and let \(v^{\max}\)
satisfy \(v_j^{\max}=\infty\) if no upper bound on reaction rate \(j\) is
known. The last reaction rate \(v_n\) is the cell growth rate. Consider the
constraints
\[
Sv=0,\qquad v\succeq0,\qquad v\preceq v^{\max}.
\]
The data are given in \texttt{fba\_data.m}.

(a) Find the maximum possible cell growth rate \(G^\star\), and optimal
Lagrange multipliers for the reaction rate limits. Determine the sensitivity of
the maximum growth rate to the reaction rate limits.
\end{quote}

\paragraph{Discussion.}
This example shows a larger cleaning operation. The biological motivation and
illustrative reaction examples are removed, while the formal variables,
constraints, data dependency, and optimization task are retained.

\subsection{Example: Dependency Recovery and Standardization}
\label{app:dep-rec-and-std-ex}

This example illustrates a case in which an exercise depends on both a target
problem reference and an algorithm reference. The algorithm further introduces a
second-level formula dependency. The recovered record therefore contains four
main components: the cleaned exercise, dependency metadata, the
context-enriched statement, and the standardized mathematical statement used
downstream.

\paragraph{Source.}
\texttt{book/Numerical Optimization\_Chp16}, Exercise 16.24.

\paragraph{Recovered dependency metadata.}
Table~\ref{tab:dep-rec-example-metadata} summarizes the dependency chain
identified during recursive dependency recovery. The original exercise directly
refers to problem (16.76) and Algorithm 16.4, while Algorithm 16.4 introduces an
additional formula dependency on (16.58).

\begin{table}[h]
\centering
\small
\caption{Recovered dependencies for Exercise 16.24.}
\label{tab:dep-rec-example-metadata}
\begin{tabular}{llll}
\toprule
Level & Field & Label & Evidence \\
\midrule
1 & \texttt{body\_refs} & \texttt{eq:16.76} & \texttt{problem (16.76)} \\
1 & \texttt{theorem\_refs} & \texttt{alg:16.4} & \texttt{refs\_attr} \\
2 & \texttt{inner\_body\_refs} & \texttt{eq:16.58} & \texttt{(16.58)} \\
2 & \texttt{expanded\_body\_refs} & \texttt{eq:16.58} & \texttt{via:Algorithm 16.4} \\
\bottomrule
\end{tabular}
\end{table}

\paragraph{Recovered record.}
The following display shows the intermediate records produced by dependency
recovery and standardization.

\subparagraph{Input. Cleaned exercise.}
\begin{quote}
\small
Exercise 16.24. Program Algorithm 16.4 and use it to solve problem (16.76).
Set all initial variables to be the vector \(e=(1,1,\ldots,1)^T\).
\end{quote}

\subparagraph{Stage 1. Context-enriched problem.}
The context-enriched version keeps the inserted references explicit and
traceable.

\medskip
\noindent\textbf{Reference formula (16.76).}
\[
\begin{array}{ll}
\max & 6x_1+4x_2-13-x_1^2-x_2^2, \\
\mbox{subject to} & x_1+x_2\leq 3,\qquad x_1\geq 0,\qquad x_2\geq 0.
\end{array}
\tag{16.76}
\]

\noindent\textbf{Reference result Algorithm 16.4.}
Algorithm 16.4, the predictor-corrector algorithm for quadratic programming,
computes \((x_0,y_0,\lambda_0)\) with \((y_0,\lambda_0)>0\). For
\(k=0,1,2,\ldots\), set
\[
(x,y,\lambda)=(x_k,y_k,\lambda_k)
\]
and solve (16.58) with \(\sigma=0\) for
\[
(\Delta x^{\mathrm{aff}},
  \Delta y^{\mathrm{aff}},
  \Delta \lambda^{\mathrm{aff}}).
\]
Then calculate
\[
\mu=\frac{y^T\lambda}{m},
\]
\[
\hat{\alpha}_{\mathrm{aff}}
=
\max\left\{
\alpha\in(0,1]
\mid
(y,\lambda)+\alpha
(\Delta y^{\mathrm{aff}},\Delta\lambda^{\mathrm{aff}})
\geq 0
\right\},
\]
and
\[
\mu_{\mathrm{aff}}
=
\frac{
(y+\hat{\alpha}^{\mathrm{aff}}\Delta y^{\mathrm{aff}})^T
(\lambda+\hat{\alpha}^{\mathrm{aff}}\Delta\lambda^{\mathrm{aff}})
}{m}.
\]
The centering parameter is then set to
\[
\sigma=\left(\frac{\mu_{\mathrm{aff}}}{\mu}\right)^3.
\]

\noindent\textbf{Second-level reference formula (16.58).}
\[
\begin{bmatrix}
G & 0 & -A^T\\
A & -I & 0\\
0 & \Lambda & \mathcal{Y}
\end{bmatrix}
\begin{bmatrix}
\Delta x\\
\Delta y\\
\Delta\lambda
\end{bmatrix}
=
\begin{bmatrix}
-r_d\\
-r_p\\
-\Lambda\mathcal{Y}e+\sigma\mu e
\end{bmatrix}.
\tag{16.58}
\]

\subparagraph{Stage 2. Standardized mathematical statement.}
The standardized version removes explicit provenance markers and rewrites the
recovered references into a self-contained mathematical task.

\begin{quote}
\small
Use the predictor-corrector algorithm for quadratic programming to solve
\[
\begin{array}{ll}
\max & 6x_1+4x_2-13-x_1^2-x_2^2, \\
\mbox{subject to} & x_1+x_2\le 3,\qquad x_1\ge 0,\qquad x_2\ge 0.
\end{array}
\]
Take all initial variables equal to \(e=(1,1,\ldots,1)^T\). Equivalently,
compute an initial point \((x_0,y_0,\lambda_0)\) with all components equal to
\(1\) and with \((y_0,\lambda_0)>0\).

For \(k=0,1,2,\ldots\), set
\[
(x,y,\lambda)=(x_k,y_k,\lambda_k)
\]
and solve, with \(\sigma=0\),
\[
\begin{bmatrix}
G & 0 & -A^T\\
A & -I & 0\\
0 & \Lambda & \mathcal{Y}
\end{bmatrix}
\begin{bmatrix}
\Delta x\\
\Delta y\\
\Delta\lambda
\end{bmatrix}
=
\begin{bmatrix}
-r_d\\
-r_p\\
-\Lambda\mathcal{Y}e+\sigma\mu e
\end{bmatrix},
\]
where
\[
r_d=Gx-A^T\lambda+c,
\qquad
r_p=Ax-y-b.
\]
This gives
\[
(\Delta x^{\mathrm{aff}},
  \Delta y^{\mathrm{aff}},
  \Delta\lambda^{\mathrm{aff}}).
\]
Next compute
\[
\mu=\frac{y^T\lambda}{m},
\]
\[
\hat\alpha_{\mathrm{aff}}
=
\max\left\{
\alpha\in(0,1]
\mid
(y,\lambda)+\alpha
(\Delta y^{\mathrm{aff}},\Delta\lambda^{\mathrm{aff}})
\ge 0
\right\},
\]
and
\[
\mu_{\mathrm{aff}}
=
\frac{
(y+\hat\alpha^{\mathrm{aff}}\Delta y^{\mathrm{aff}})^T
(\lambda+\hat\alpha^{\mathrm{aff}}\Delta\lambda^{\mathrm{aff}})
}{m}.
\]
Set the centering parameter to
\[
\sigma=\left(\frac{\mu_{\mathrm{aff}}}{\mu}\right)^3.
\]
Solve the given quadratic programming problem with this initialization.
\end{quote}

\paragraph{Discussion.}
This example highlights the distinction between enrichment and normalization.
The context-enriched record preserves the provenance of inserted references in a
traceable form, while the standardized mathematical statement converts those
recovered dependencies into a cleaner self-contained problem for downstream
completion and review.
\subsection{Example: Self-Contained Target Construction}
\label{app:enriched_to_self-contained_exp}

This example illustrates how a standardized mathematical statement is converted
into the final self-contained informal target. The final target makes domains,
definitions, and implicit well-formedness conditions explicit before Lean
formalization.

\paragraph{Input. Cleaned exercise.}
\begin{quote}
Exercise 8.14. Let $C$ be a polyhedron in $\mathbf{R}^n$ described as
$C = \{x \mid Ax \preceq b\}$, and suppose that
$\{x \mid Ax \prec b\}$ is nonempty. (a) Show that the maximum volume ellipsoid
enclosed in $C$, expanded by a factor $n$ about its center, contains $C$.
\end{quote}

\paragraph{Stage 1. Standardized mathematical statement.}
\begin{quote}
Let
\[
C=\{x\in \mathbf{R}^n\mid Ax\preceq b\},
\]
and assume that the strict inequality system is feasible, i.e.,
\[
\{x\in \mathbf{R}^n\mid Ax\prec b\}\neq \emptyset.
\]
Show that if an ellipsoid of maximum volume is contained in \(C\), then the
ellipsoid obtained by expanding it by a factor \(n\) about its center contains
\(C\).
\end{quote}

\paragraph{Stage 2. Self-contained target.}
\begin{quote}
Let \(A \in \mathbf{R}^{m \times n}\) and \(b \in \mathbf{R}^m\), and define
\[
C=\{x\in \mathbf{R}^n \mid Ax \preceq b\},
\]
where \((Ax)_i \le b_i\) for all \(i=1,\dots,m\). Assume
\[
\{x\in \mathbf{R}^n \mid (Ax)_i < b_i
\text{ for all } i=1,\dots,m\}\ne \emptyset.
\]
An ellipsoid in \(\mathbf{R}^n\) is a set of the form
\[
E=\{x\in \mathbf{R}^n \mid (x-c)^T P^{-1}(x-c)\le 1\},
\]
where \(c\in \mathbf{R}^n\) and \(P\in \mathbf{R}^{n\times n}\) is symmetric
positive definite. For \(\alpha>0\), define the expansion of \(E\) about its
center \(c\) by
\[
c+\alpha(E-c)=\{c+\alpha(x-c)\mid x\in E\}.
\]
Show that if \(E\) is a maximum-volume ellipsoid contained in \(C\), then
\(c+n(E-c)\) contains \(C\).
\end{quote}

\paragraph{Discussion.}
The final target preserves the original mathematical task while making the
objects needed for a formal theorem statement explicit. In particular, it
specifies the dimensions of \(A\) and \(b\), expands the componentwise
inequality notation, defines the ellipsoid model, and formalizes the expansion
operation about the center.

\subsection{Example: Multi-Block Decomposition}
\label{app:example-block-decomposition}

We illustrate the block decomposition
(Section~\ref{app:semantic-block-decomposition})
with Exercise~3.12 from the Boyd--Vandenberghe extra exercises,
which concerns a trace-minimization semidefinite program (SDP).
The preprocessor decomposes the exercise into four ordered records:

\begin{center}
\small
\begin{tabular}{clll}
\toprule
\textbf{index} & \textbf{kind} & \textbf{term} & \textbf{content (abbreviated)} \\
\midrule
4 & \texttt{defn}      & semidefinite program
  & A semidefinite program has a symmetric-matrix variable\ldots \\
5 & \texttt{opt\_prob} & trace minimization SDP
  & minimize $\operatorname{tr} X$ s.t.\ $\begin{bmatrix} A & B \\ B^\top & X \end{bmatrix} \succeq 0$ \\
6 & \texttt{thm}       & --- 
  & Show $X = B^\top A^{-1}B$ solves the SDP. \\
7 & \texttt{thm}       & --- 
  & Conclude $(A,B)\mapsto \operatorname{tr}(B^\top A^{-1}B)$ is convex. \\
\bottomrule
\end{tabular}
\end{center}

The four records are translated according to their semantic roles.
Block~4 (\texttt{defn}) is retained as provenance and contextual
information, but it does not necessarily produce a standalone Lean declaration.
Block~5 (\texttt{opt\_prob}) is translated into Lean declarations that encode
the problem data, the feasibility predicate, and the objective function.
In the generated Lean file, these declarations appear under explicit
block-boundary comments:

\begin{lstlisting}[language=lean]
/- [BLOCK Exercise 3.12 | 5 | opt_prob] -/
structure TraceMinimizationSDP (m n : ℕ) where
  A        : Matrix (Fin m) (Fin m) ℝ
  A_symm   : Aᵀ = A
  A_posDef : Matrix.PosDef A
  B        : Matrix (Fin m) (Fin n) ℝ
\end{lstlisting}
\vspace{1.5cm}
\begin{lstlisting}[language=lean]
def TraceMinimizationSDP.isFeasible {m n : ℕ}
    (P : TraceMinimizationSDP m n)
    (X : Matrix (Fin n) (Fin n) ℝ) : Prop :=
  Xᵀ = X ∧ Matrix.PosSemidef (P.blockMatrix X)

def TraceMinimizationSDP.traceObjective {m n : ℕ}
    (P : TraceMinimizationSDP m n)
    (X : Matrix (Fin n) (Fin n) ℝ) : ℝ :=
  Matrix.trace X
\end{lstlisting}

Blocks~6 and~7 become separate Lean theorem targets with
\texttt{by sorry} proof bodies. These proof placeholders are intentional:
they mark the targets that are later passed to downstream proof generation
and verification.
\vspace{0.2cm}
\begin{lstlisting}[language=lean]
/- [BLOCK Exercise 3.12 | 6 | thm] -/
theorem traceMinimizationSDP_solution
    {m n : ℕ} (P : TraceMinimizationSDP m n) :
    P.isFeasible (P.Bᵀ * P.A⁻¹ * P.B) ∧
      ∀ X : Matrix (Fin n) (Fin n) ℝ, P.isFeasible X →
        Matrix.trace (P.Bᵀ * P.A⁻¹ * P.B) ≤ Matrix.trace X := by
  sorry

/- [BLOCK Exercise 3.12 | 7 | thm] -/
theorem trace_btranspose_inv_mul_b_convex_on {m n : ℕ} :
    ConvexOn ℝ
      {p : Matrix (Fin m) (Fin m) ℝ × Matrix (Fin m) (Fin n) ℝ |
         p.1ᵀ = p.1 ∧ Matrix.PosDef p.1}
      (fun p => Matrix.trace (p.2ᵀ * p.1⁻¹ * p.2)) := by
  sorry
\end{lstlisting}

The block-boundary comments, such as
\texttt{/- [BLOCK Exercise 3.12 | 6 | thm] -/}, are inserted by the
orchestrator before each generated declaration. They make the Lean artifact
traceable back to the corresponding semantic record.

\subsection{Example: One Translation--Repair Cycle}
\label{app:example-compile-repair}

We illustrate the compilation-repair loop
(Section~\ref{app:blockwise-construction-repair})
with the \texttt{thm} block for Exercise~3.7-(b), which asks to show that
a power-constraint set is convex. The translator's first attempt used an
invalid summation form for the target Lean~4 environment:
\begin{lstlisting}[language=lean]
-- First attempt: compile fails.
theorem power_constraint_set_convex
    (m n : ℕ) (hm : 10 ≤ m) (a : Fin n → Fin m → ℝ)
    (I_des : ℝ) (hI : 0 < I_des) :
    Convex ℝ {p : Fin m → ℝ |
      (∀ j, 0 ≤ p j) ∧
      ∀ S : Finset (Fin m), S.card = 10 →
        ∑ j in S, p j ≤ (1 / 2 : ℝ) * ∑ j : Fin m, p j} := by
  sorry
\end{lstlisting}

Lean emitted the following diagnostic:

\begin{lstlisting}[language=lean]
unexpected token 'in'; expected ','
  ∑ j in S, p j ≤ (1 / 2 : ℝ) * ∑ j : Fin m, p j
\end{lstlisting}

The repair step receives this diagnostic together with the frozen Lean prefix.
It then rewrites only the current block, replacing the unsupported notation with
a Lean~4-compatible expression:

\begin{lstlisting}[language=lean]
-- Repair r = 1: compile succeeds.
theorem power_constraint_set_convex
    (m n : ℕ) (hm : 10 ≤ m) (a : Fin n → Fin m → ℝ)
    (I_des : ℝ) (hI : 0 < I_des) :
\end{lstlisting}
\vspace{1cm}
\begin{lstlisting}[language=lean]
    Convex ℝ {p : Fin m → ℝ |
      (∀ j, 0 ≤ p j) ∧
      ∀ S : Finset (Fin m), S.card = 10 →
        Finset.sum S p ≤ (1 / 2 : ℝ) * ∑ j : Fin m, p j} := by
  sorry
\end{lstlisting}

The combined Lean file compiles after one repair attempt, so the corrected
block is accepted and frozen. Later blocks can use it as read-only context, but
the repair loop is not allowed to modify it again.

\subsection{Example: One Semantic Review--Rewrite Cycle}
\label{app:example-semantic-repair}

We illustrate the semantic review loop
(Section~\ref{app:semantic-repair-loop})
with the \texttt{thm} block for Exercise~9.4-(a) in
\emph{Numerical Optimization}. The exercise asks to show that six distinct
collinear points in $\mathbf{R}^2$ do not uniquely determine a quadratic.
The initial translation produced a Lean statement that compiled, but its
hypothesis did not faithfully encode collinearity:

\begin{lstlisting}[language=lean]
-- Compilable but semantically incorrect.
theorem six_collinear_points_do_not_determine_quadratic_uniquely
    (P : Fin 6 → ℝ × ℝ)
    (hcol : ∃ A B C : ℝ,
      A ≠ 0 ∨ B ≠ 0 ∧
      ∀ i : Fin 6, A * (P i).1 + B * (P i).2 = C)
    (hdistinct : Function.Injective P) :
    ∃ q₁ q₂ : ℝ × ℝ → ℝ, q₁ ≠ q₂ ∧
      q₁ ≠ q₂ ∧
      ∀ i : Fin 6, q₁ (P i) = q₂ (P i) := by
  sorry
\end{lstlisting}

Because \texttt{\(\wedge\)} binds more tightly than \texttt{\(\vee\)}, the
hypothesis \texttt{hcol} is parsed as
\[
\exists A\,B\,C,\;
A \neq 0 \;\vee\;
\bigl(B \neq 0 \wedge
\forall i,\; A(P_i)_1 + B(P_i)_2 = C\bigr).
\]

Thus, in the branch \texttt{A $\neq$ 0}, the theorem imposes no line constraint on
the six points. The semantic reviewer therefore flags a
\texttt{missing\_assumption} issue:

\begin{lstlisting}
{
  "math_equivalent": false,
  "overall_status": "not_usable_yet",
  "dominant_issue_types": ["missing_assumption"],
  "top_priority_fix":
    "Fix logical grouping in hcol: require (A ≠ 0 ∨ B ≠ 0)
     and the pointwise line condition together.",
  "issues": [{
    "severity": "P0",
    "issue_type": "missing_assumption",
    "reason":
      "In the A ≠ 0 branch, no condition is imposed on the points,
       so collinearity is not actually assumed.",
    "suggested_fix":
      "∃ A B C : ℝ, (A ≠ 0 ∨ B ≠ 0) ∧
       ∀ i : Fin 6, A * (P i).1 + B * (P i).2 = C"
  }]
}
\end{lstlisting}

The semantic rewrite step corrects the logical grouping while preserving the
surrounding theorem structure:

\begin{lstlisting}[language=lean]
-- After semantic rewrite: compile succeeds and reviewer accepts.
theorem six_collinear_points_do_not_determine_quadratic_uniquely
    (P : Fin 6 → ℝ × ℝ)
    (hcol : ∃ A B C : ℝ,
      (A ≠ 0 ∨ B ≠ 0) ∧
      ∀ i : Fin 6, A * (P i).1 + B * (P i).2 = C)
\end{lstlisting}
\vspace{1cm}
\begin{lstlisting}[language=lean]
    (hdistinct : Function.Injective P) :
    ∃ q₁ q₂ : ℝ × ℝ → ℝ, q₁ ≠ q₂ ∧
      q₁ ≠ q₂ ∧
      ∀ i : Fin 6, q₁ (P i) = q₂ (P i) := by
  sorry
\end{lstlisting}

The rewritten statement compiles immediately, so no additional
compilation-repair sub-loop is triggered. This example shows why semantic review
is needed even after Lean accepts the file: a precedence error can silently
weaken a mathematical assumption without causing a compiler diagnostic.

\subsection{Example: Source-Level Review}
\label{app:source-level-review-exp}

This example shows representative source-level review records. Each record
compares the final informal target against the original problem and assigns a
review label such as \textbf{accept}, \textbf{revise}, or \textbf{hold}. The
records are shown in their original JSON-style format.

\begin{jsonbox}
[
  {
    "case": "accept",
    "source_idx": "Exercise 7.2",
    "source": "book/Numerical Optimization_Chp7",
    "problem": "Show that the matrix $\hat{H}_{k+1}$ in (7.21) is singular.",
    "problem_finally": "Let $f:\mathbb{R}^n \to \mathbb{R}$ be differentiable, and let $s_k,y_k,p_k \in \mathbb{R}^n$ satisfy $y_k^T s_k \neq 0$. Define $\hat{H}_{k+1}=I - s_k y_k^T/(y_k^T s_k)$, where $I$ is the $n \times n$ identity matrix. Suppose $p_{k+1}=...=-\hat{H}_{k+1} \nabla f_{k+1}$. Show that $\hat{H}_{k+1}$ is singular.",
    "semantic_report": {
      "overall_status": "accept",
      "reason": "The final version preserves the original goal, makes the needed nonzero denominator assumption explicit, is self-contained, and remains theorem-style. Boundary case check: even for $n=1$, $\hat{H}_{k+1}=1-(s_k y_k)/(y_k s_k)=0$, so singularity still holds.",
      "issues": []
    }
  },
  {
    "case": "revise / missing_assumption",
    "source_idx": "Exercise 14.8",
    "source": "book/Numerical Optimization_Chp14",
    "problem": "Prove that the coefficient matrix in (14.16) is nonsingular if and only if $A$ has full row rank.",
    "problem_finally": "Let $m,n \in \mathbb{N}$, let $A \in \mathbb{R}^{m \times n}$, and let $X,S \in \mathbb{R}^{n \times n}$ be diagonal matrices. Prove that the block matrix $[[0,A^T,I],[A,0,0],[S,0,X]]$ is nonsingular if and only if $A$ has full row rank.",
    "semantic_report": {
      "overall_status": "revise",
      "reason": "The final statement drops the interior-point context assumption that $X$ and $S$ have positive/nonzero diagonal entries. As written it is false: if $A$ has full row rank but $X=0$ and $S=0$, the block matrix is singular.",
      "true_issue": "missing_assumption",
      "suggestion_fix": "Add that $X$ and $S$ are diagonal matrices with strictly positive diagonal entries, equivalently invertible diagonal matrices."
    }
  },
  {
    "case": "revise / task_drift",
    "source_idx": "Exercise 7.7",
    "source": "book/Numerical Optimization_Chp7",
    "problem": "Minimum surface problem. Divide each edge of the unit square into $q$ intervals, label the $(q+1)^2$ grid points row by row, approximate derivatives by finite differences, and show that $f_j$ has the displayed form involving $(x_j-x_{j+q+1})^2$ and $(x_{j+1}-x_{j+q})^2$.",
    "problem_finally": "Let $q \ge 1$. Divide $[0,1]^2$ into $q \times q$ subsquares. Label the $(q+1)^2$ grid points row by row. For each subsquare $A_j$, indexed by its lower-left vertex, let its four vertices correspond to $x_j$, $x_{j+1}$, $x_{j+q}$, $x_{j+q+1}$. Show the approximate area contribution formula and determine the interior variables minimizing $\sum_j f_j(x)$.",
    "semantic_report": {
      "overall_status": "revise",
      "reason": "The final version changes the subsquare vertex indexing. With row-by-row indexing on a grid of width $q+1$, the vertical offset should be $q+1$, not $q$. This is a substantive mathematical drift and conflicts with the formula.",
      "true_issue": "task_drift",
      "suggestion_fix": "Use vertices $x_j$, $x_{j+1}$, $x_{j+q+1}$, $x_{j+q+2}$ consistently, or otherwise repair the indexing."
    }
  },
  {
    "case": "hold / missing_dependency",
    "source_idx": "Exercise 7.3",
    "source": "book/Numerical Optimization_Chp7",
    "problem": "Derive the formula (7.23) under the assumption that line searches are exact.",
    "problem_finally": "Let $f:\mathbb{R}^n \to \mathbb{R}$ be differentiable. For each iteration $k$, let $x_{k+1}$, $p_k$, $y_k$, and $H_{k+1}$ be such that the expressions are defined. Assume exact line search, so $\nabla f(x_{k+1})^T p_k = 0$, and assume $y_k^T p_k \neq 0$. Derive $p_{k+1}=-H_{k+1} \nabla f(x_{k+1}) = -\nabla f(x_{k+1}) + ... p_k$.",
    "semantic_report": {
      "overall_status": "hold",
      "hold_reason_type": "missing_dependency",
      "reason": "The statement asks to derive formula (7.23) but does not give the defining relation for $H_{k+1}$ or the algorithmic update context. Thus the conclusion does not follow from the stated assumptions.",
      "suggestion_fix": "Restate the missing update relation or prior formula defining $H_{k+1}$ and $p_{k+1}$."
    }
  },
  {
    "case": "hold / unsuitable_lean",
    "source_idx": "Exercise 18.5",
    "source": "book/Numerical Optimization_Chp18",
    "problem": "Consider the constraint $x_1^2+x_2^2=1$. Write the linearized constraints (18.7b) at the points $(0,0)^T$, $(0,1)^T$, $(0.1,0.02)^T$, and $-(0.1,0.02)^T$.",
    "problem_finally": "Let $c:\mathbb{R}^2 \to \mathbb{R}$ be $c(x)=x_1^2+x_2^2-1$. For a point $x_k$, define $c_k=c(x_k)$ and $A_k=\nabla c(x_k)^T$. The linearized equality constraint is $A_k p + c_k = 0$. Write this constraint explicitly at the four listed points.",
    "semantic_report": {
      "overall_status": "hold",
      "hold_reason_type": "unsuitable_lean",
      "reason": "The statement is self-contained and faithful, but the task is phrased as a compute/write-out exercise rather than a theorem-style proposition suitable for Lean formalization.",
      "suggestion_fix": "Rewrite as explicit theorem statements giving the four resulting linearized constraints."
    }
  }
]
\end{jsonbox}

\paragraph{Discussion.}
These examples show how the semantic reviewer separates faithful, self-contained
targets from cases requiring repair or exclusion. The \textbf{revise} cases are
locally repairable, while the \textbf{hold} cases require missing dependencies
or task-form issues to be resolved before formalization.

\subsection{Example: Constrained Repair and Re-review}
\label{app:repair_and_re-review_exp}

This example shows two representative repair cases. The first is a local
assumption-restoration case that remains in the \texttt{revise} path. The
second begins as a \texttt{hold} case because the task form is not yet suitable
for Lean and a missing definition issue must also be resolved before a precise
theorem-style target can be stated.

\subsubsection{Case 1: Revise / Missing Assumption}

\begin{promptbox}
\small
{\ttfamily \{}\par
{\ttfamily \quad "source\_idx":} "Exercise 14.8",\par
{\ttfamily \quad "repair\_type":} "revise / missing\_assumption",\par
{\ttfamily \quad "original\_problem":} "Exercise 14.8. Prove that the coefficient matrix in (14.16) is nonsingular if and only if $A$ has full row rank.",\par
{\ttfamily \quad "before\_repair\_problem\_finally":} "Let $m,n \in \mathbf{N}$, let $A \in \mathbf{R}^{m \times n}$, and let $X,S \in \mathbf{R}^{n \times n}$ be diagonal matrices. Prove that the block matrix
\[
\begin{bmatrix}
0 & A^T & I \\
A & 0 & 0 \\
S & 0 & X
\end{bmatrix}
\]
is nonsingular if and only if $A$ has full row rank.",\par
{\ttfamily \quad "repair\_reason":} "The original context needs $X$ and $S$ to be positive/invertible diagonal matrices. Without this, e.g. $X=0$ and $S=0$ gives a singular block matrix even when $A$ has full row rank.",\par
{\ttfamily \quad "after\_repair\_problem\_finally":} "Let $m,n \in \mathbf{N}$, let $A \in \mathbf{R}^{m \times n}$, and let $X,S \in \mathbf{R}^{n \times n}$ be diagonal matrices with strictly positive diagonal entries. Prove that the block matrix
\[
\begin{bmatrix}
0 & A^T & I \\
A & 0 & 0 \\
S & 0 & X
\end{bmatrix}
\]
is nonsingular if and only if $A$ has full row rank."\par
{\ttfamily \}}
\end{promptbox}

\subsubsection{Case 2: Hold / Unsuitable Lean and Missing Definition}

\begin{promptbox}
\small
{\ttfamily \{}\par
{\ttfamily \quad "source\_idx":} "Exercise 7.4",\par
{\ttfamily \quad "repair\_type":} "hold / unsuitable\_lean + missing\_def",\par
{\ttfamily \quad "original\_problem":} "Exercise 7.4. Consider limited-memory SR1 updating based on (7.30). Explain how the storage can be cut in half if the basic matrix $B_0^k$ is kept fixed for all $k$. (Hint: Consider the matrix $Q_k=[q_0,\ldots,q_{k-1}]=Y_k-B_0S_k$.)",\par
{\ttfamily \quad "before\_repair\_problem\_finally":} "Consider the limited-memory SR1 update with memory parameter $m \in \mathbf{N}$. For iteration $k$, let $p=\min\{k,m\}$, and let the stored quasi-Newton pairs be $s_\ell,y_\ell \in \mathbf{R}^n$. Define $S_k=[\,s_{k-m},\ldots,s_{k-1}\,]\in \mathbf{R}^{n\times p}$ and $Y_k=[\,y_{k-m},\ldots,y_{k-1}\,]\in \mathbf{R}^{n\times p}$, and define the strictly lower-triangular matrix $L_k$ by
\[
(L_k)_{i,j}=\begin{cases}
(s_{k-m-1+i})^T y_{k-m-1+j}, & i>j,\\
0, & i\le j.
\end{cases}
\]
Assume that the basic matrix is fixed for all $k$, so $B_0^k=B_0\in \mathbf{R}^{n\times n}$ for every $k$. Define $Q_k=[q_0,\ldots,q_{p-1}]=Y_k-B_0S_k$, with $q_i=y_i-B_0s_i$ for each stored pair. Explain how the storage can be reduced by one-half under the assumption that $B_0^k$ is fixed for all $k$.",\par
{\ttfamily \quad "repair\_reason":} "Before repair, the statement remains open-ended because it asks the reader to explain how the storage reduction works, which is not suitable as a Lean theorem target. In addition, the column indices $q_i$ of $Q_k$ are inconsistent with the storage-window indexing used for $S_k$ and $Y_k$.",\par
{\ttfamily \quad "after\_repair\_problem\_finally":} "Let $m \in \mathbf{N}$, let $k \in \mathbf{N}$, and let $p=\min\{k,m\}$. For the stored quasi-Newton pairs $(s_j,y_j) \in \mathbf{R}^n \times \mathbf{R}^n$, define $S_k=[s_{k-p},\ldots,s_{k-1}] \in \mathbf{R}^{n\times p}$ and $Y_k=[y_{k-p},\ldots,y_{k-1}] \in \mathbf{R}^{n\times p}$. Assume the basic matrix is fixed, i.e. $B_0^k=B_0 \in \mathbf{R}^{n\times n}$ for all $k$. For $i=0,\ldots,p-1$, define
\[
q_i := y_{k-p+i}-B_0 s_{k-p+i},
\]
and let
\[
Q_k=[q_0,\ldots,q_{p-1}] = Y_k-B_0S_k.
\]
Define the strictly lower-triangular matrix $L_k\in \mathbf{R}^{p\times p}$ by
\[
(L_k)_{i,j}=\begin{cases}
s_{k-p+i}^T y_{k-p+j}, & i>j,\\
0, & i\le j,
\end{cases}
\qquad (i,j=0,\ldots,p-1).
\]
Prove that for all $i>j$,
\[
(L_k)_{i,j}=s_{k-p+i}^T q_j + s_{k-p+i}^T B_0 s_{k-p+j}.
\]
Hence, when $B_0$ is fixed, the information needed to form $L_k$ can be recovered from $S_k$, $Q_k$, and $B_0$, so it is unnecessary to store both $Y_k$ and $Q_k$; equivalently, one may replace storage of the pair $(S_k,Y_k)$ by storage of $(S_k,Q_k)$, which cuts the storage for the update pairs in half."\par
{\ttfamily \}}
\end{promptbox}

\paragraph{Discussion.}
These two cases illustrate the intended boundary of constrained repair. In the
first case, the reviewer identifies a single missing assumption, and the repair
restores the positivity condition needed for the original equivalence to be
correct. In the second case, the original target is not merely incomplete but
also unsuitable for direct formalization: the open-ended instruction must be
converted into a precise theorem-style statement, and the indexing of the stored
quantities must be aligned before re-review can succeed.

\subsection{Examples: Representative formalization failures}
\label{app:formalization_failures}
\textbf{Library/API grounding.}
   For example, in a convexity target about monotonicity of gradients, the final
   snapshot contains:
\vspace{-0.5em}
\begin{lstlisting}[language=Lean,basicstyle=\ttfamily\footnotesize]
theorem gradient_monotone_of_convex_differentiable
    {n : ℕ}
    {f : EuclideanSpace ℝ (Fin n) → ℝ}
    (hconv : ConvexOn ℝ Set.univ f)
    (hfdiff : Differentiable ℝ f) :
    MonotoneMap (fun x => gradient f x) := by
  intro x y
  simpa [gradient, sub_eq_add_neg, inner_add_left, inner_add_right,
    inner_neg_left, inner_neg_right]
    using hconv.monotoneOn_gradient hfdiff x (by simp) y (by simp)
\end{lstlisting}

The intended mathematical route is reasonable: convex differentiable functions
have monotone gradients. However, the proof is not grounded in actual
Lean/Mathlib declarations. The compiler reports an invalid-field error for
\texttt{monotoneOn\_gradient}.

\textbf{Formal representation and type discipline.}
Matrix-analysis targets often expose the gap between informal linear algebra
notation and Lean's exact representation of dimensions, indices, rank, and
positive-definiteness predicates.
\begin{lstlisting}[language=Lean,basicstyle=\ttfamily\footnotesize]
theorem full_column_rank_iff_transpose_mul_self_posDef
    {m n : Type*} [Fintype m] [Fintype n] [DecidableEq m] [DecidableEq n]
    (h : Fintype.card n ≤ Fintype.card m) (J : Matrix m n ℝ) :
    J.rank = Fintype.card n ↔
      Matrix.PosDef (J.transpose * J) := by
  simpa [Matrix.posDef_iff]
    using Matrix.rank_transpose_mul_self_eq_card_iff_posDef
      (m := m) (n := n) J
\end{lstlisting}

This proof mirrors the standard informal theorem that full column rank is
equivalent to positive definiteness of \(J^\top J\). The Lean attempt fails
because the required representation bridge is not available under these names:
the compiler reports unknown constants such as
\texttt{Matrix.posDef\_iff} and
\texttt{Matrix.rank\_transpose\_mul\_self\_eq\_card\_iff\_posDef}. The failure is
therefore not just a missing algebraic idea; it reflects the need to align the
informal matrix theorem with Mathlib's actual rank API, matrix index types, and
positive-definiteness definitions.

\textbf{Proof decomposition and long-horizon control.}
Agentic systems also fail when a proof requires a long chain of auxiliary
analytic lemmas. In an Aristotle run on \texttt{problem-77.lean}, the target is
Q-linear convergence of Newton's method at a multiple root:

\begin{lstlisting}[language=Lean,basicstyle=\ttfamily\footnotesize]
theorem classicalNewtonIteration_multipleRoot_qLinear
    (N : ClassicalNewtonIteration)
    (hroot : IsMultipleRoot N.f 0)
    (hf'_correct : ∀ x : ℝ, N.f' x = deriv N.f x)
    (hC1 : ∃ s : Set ℝ, s ∈ �� 0 ∧ ContDiffOn ℝ 1 N.f s) :
    IsQLinearlyConvergentTo N.xSeq 0 := by
  sorry
\end{lstlisting}

The run did make structured progress: its summary reports helper lemmas such
as \texttt{f\_zero\_of\_multipleRoot},
\texttt{hasDerivAt\_zero\_of\_multipleRoot},
\texttt{seq\_ne\_zero}, \texttt{newton\_step\_eq}, and
\texttt{newton\_contraction\_ratio}. The remaining gap was the analytic lemma
\texttt{newton\_ratio\_tendsto}, which must show that
\[
  x_k^{m-1} g(x_k) / f'(x_k) \to 1/m
\]
along the Newton sequence. This is not a local syntax error: closing the proof
requires maintaining a convergence argument through multiple helper lemmas,
derivative facts, and asymptotic estimates. It is representative of
long-horizon proof-control failures on optimization and numerical-analysis
targets.

\section{Prompt Templates}
\label{app:prompt-templates}

This appendix lists the prompt templates used in the benchmark construction and
evaluation pipeline. The prompts are organized in the same order as the pipeline:
raw exercise cleaning, self-contained target construction, source-level review,
constrained repair, JSON-to-Lean translation, and pass@$k$ proof generation.

\subsection{Raw-to-Clean Prompt}
\label{app:raw_to_clean_prompt}

This prompt is used to construct the \texttt{problem\_clean} field from the raw
extracted exercise. Its purpose is to remove non-essential hints, remarks,
background narrative, and implementation-oriented text while preserving the
mathematical statement and all meaningful references.

\paragraph{Prompt template.}
\begin{promptbox1}
You are a preprocessing cleaner for optimization and mathematics exercise statements.

Role.
Your task is to remove hint/remark prompt-like information and background narrative from the original problem statement, while keeping all valid mathematical information unchanged, keeping all tags unchanged, and cleaning the original problem text only.

Input.
A single JSON object with a field problem containing raw exercise text.

Rules.
1. Preserve the mathematical meaning exactly.
2. Preserve symbols, formulas, numbering, and mathematically meaningful references.
3. Do not delete, alter, renumber, or rewrite any referenced identifier or index, including equation numbers, theorem numbers, algorithm numbers, exercise numbers, part labels, and cross-references such as "as in (4.13)", "part (b)", "Theorem 12.1", or "Algorithm 3.2".
4. Preserve referential links exactly: if a clause refers to an earlier object by number or label, keep that number or label unchanged and attached to the same object.
5. Do not change the task intent, target, or scope. For example, do not convert "prove" to "compute", do not replace "minimum and maximum" with only "minimum", and do not weaken or strengthen quantifiers or assumptions.
6. Cleaning only: remove noise, wrappers, and non-mathematical exposition, but do not rewrite core mathematical content into a different statement.
7. Do not introduce new claims, examples, explanations, or solution steps.
8. Mandatory boundary-case safeguard: before treating a cleaned statement as acceptable, explicitly check edge cases such as nonemptiness and zero-valued conditions. If an edge case fails, construct a concrete counterexample and treat the item as needing revise with a clear reason in downstream review.

You must always produce a cleaned version that is shorter or at least meaningfully rewritten compared to the input. Never return the input text unchanged.

Primary objective.
Keep only mathematically essential content. Delete all irrelevant exposition. The cleaned result should read like a compact mathematical problem statement, not like textbook narration, software instructions, or pedagogical commentary.

Lean-oriented principle.
Treat the input as material to be normalized before formalization.
Prefer content that can later become:
1. definitions;
2. hypotheses;
3. variables and domains;
4. assumptions and constraints;
5. equations and inequalities;
6. precise goals and conclusions.

Delete anything that does not help identify or formalize these items. Default to deletion.

Keep only the following content:
1. exercise numbers, part labels, and mathematically meaningful cross-references such as 4.13, (a), Theorem X.Y, Algorithm X.Y, equation (15), or part (b);
2. definitions, notation, symbol declarations, and domain declarations;
3. assumptions, hypotheses, regularity conditions, feasibility assumptions, and rank or positivity assumptions;
4. optimization variables, objectives, constraints, equalities, inequalities, quantifiers, and formulas;
5. dual problems, KKT conditions, conjugates, SDP/QP/SOCP/LP formulations, feasibility claims, and exact statements of derived reformulations when they are part of the problem;
6. explicit problem data needed for the statement, including matrices, vectors, dimensions, constants, tolerances, and named data files when required to define the task;
7. the actual task directives, such as show, prove, derive, find, compute, determine, formulate, verify, construct, characterize, or solve;
8. minimal clarifying text only if removing it would destroy mathematical meaning.

Delete aggressively:
1. motivational background, applications, interpretation, intuition, historical remarks, and storytelling, including introductory sentences that describe why a topic is useful or interesting;
2. audience-directed wording such as "you may", "you will", "you should", "you need", "your job is", "we seek", "we consider", "we are interested in", "as a courtesy", "answer the following", "you should feel free to", and similar phrases;
3. all Hint, Hints, Remark, Remarks, Note, and Notes sections as labeled discourse, including multi-language hint blocks such as Matlab/Python/Julia subsections inside a Hints section;
4. solution strategies, method suggestions, and pedagogical advice, including phrases such as "you can do this by", "one approach is to", "a standard method", or "a typical method";
5. explanatory prose that merely paraphrases formulas already present;
6. software-tutorial language such as "Hello World", "how to use CVX", "tricks for using CVX", or similar presentation wording;
7. MATLAB, Python, Julia, CVX, CVXPY, Convex.jl, solver, and implementation instructions, except for mathematical data or tolerances that are part of the problem statement;
8. code snippets, random-seed commands, API usage directions, and multi-language code blocks, unless the exact numeric instance defined there is essential problem data;
9. comments such as "it is easy to see", "roughly speaking", "you can think of", "in a few very simple cases", "evidently", "of course", and similar non-mathematical narration;
10. all plotting, graphing, figure-generation, and visualization instructions, such as "Plot X vs Y", "Generate a figure", or "Graph the trajectory". These are presentation tasks, not mathematical content for formalization;
11. comparison-to-baseline instructions such as "compare it to the non-hybrid version" or "compare with the naive method", unless the comparison itself is a mathematical claim to prove;
12. "Briefly explain why" and "explain your reasoning" wrappers. Keep only the underlying mathematical task, such as "Determine whether..." or "Show that...".

Mandatory treatment of Hint and Remark.
Delete the labels Hint, Hints, Remark, Remarks, Note, and Notes from the output.
Do not preserve hint or remark paragraphs as paragraphs.
If a Hint, Remark, or Note contains indispensable mathematical content, such as a formula, inequality, or fact needed to state the problem, extract only that mathematical content and merge it into the cleaned problem statement without the discourse wrapper.
If a Hint or Remark section consists entirely of software instructions, such as MATLAB code, Python snippets, Julia variable declarations, or CVX API directions, delete the entire section with no extraction.
\end{promptbox1}

\subsection{Self-Contained Target Construction Prompt}
\label{app:enriched_to_self-contained_prompt}

This prompt is used to construct \texttt{problem\_finally} from
\texttt{problem\_standardized\_math}, with \texttt{problem\_with\_context} used
only for explicitly recovered dependencies. Its purpose is to produce a
condition-complete, self-contained informal target before Lean formalization.

\paragraph{Prompt template.}
\begin{promptbox1}
You are given one optimization exercise statement.
Rewrite it into a condition-complete, self-contained, textbook-style problem.

Input use rule.
- Primary source: problem_standardized_math. Preserve its task intent, core assumptions, and structure first.
- Secondary source: problem_with_context. Use it only to supplement missing dependency assumptions or formulas, including Reference, Double, Third, or More reference lines, when clearly relevant.
- Before all other processing, ignore any hint/hints content entirely; hint text is disposable and must not be used in rewriting.
- If problem_standardized_math and problem_with_context differ, prefer problem_standardized_math unless context provides explicit supporting mathematics needed for completeness.
- If problem_standardized_math lacks necessary mathematical support, such as assumptions, formulas, or definitions, explicitly retrieve and integrate that support from problem_with_context.
- Keep strict dependency folding: third-hop to second-hop, second-hop to first-hop, and first-hop to the original problem. If more-level dependencies exist, unfold and fold them layer-by-layer before final rewrite.
- Carefully check that every concept, symbol, function, and operator used in the problem is defined. If any definitions are missing, add precise definitions so the problem is fully self-contained and suitable for formalization.

Hard constraints.
- Do not solve the problem.
- Do not use, preserve, paraphrase, or output any hint/hints text; discard hints directly.
- Do not rewrite, split, reorder, or restate the original task objective in any form.
- The task objective must remain strictly identical to problem_standardized_math.
- Only identify and explicitly add implicit assumptions or conditions that are directly supported by input context.
- Add only information that is genuinely useful for mathematical completeness, dependency resolution, or precise definition; do not add irrelevant background, commentary, examples, or decorative exposition.
- Keep the original mathematical intent and task type.
- Do not add unnecessary stronger assumptions.
\end{promptbox1}

\subsection{Source-Level Review Prompt}
\label{app:source-level-review-prompt}

This prompt is used to review each \texttt{problem\_finally} against the
original \texttt{problem}. The reviewer checks mathematical faithfulness,
self-containedness, missing assumptions, unresolved dependencies, and suitability
for Lean formalization.

\paragraph{Prompt template.}
\begin{promptbox1}
Role.
You are reviewing a mathematical exercise for Lean formalization readiness.

Your task is to determine whether the candidate final version preserves the original mathematical meaning, includes all necessary assumptions, is fully self-contained, and is suitable for Lean formalization.

You are not asked to improve wording for style alone. Focus on mathematical faithfulness, completeness, self-containedness, and formalizability.

Before deciding accept, you must explicitly check boundary cases such as nonemptiness and zero-valued edge conditions. If a boundary case fails, provide a concrete counterexample.

Core objective.
Compare problem_finally against problem.

Your review must answer four questions:
1. Does problem_finally preserve all assumptions from the original problem?
2. Does problem_finally preserve the original task and conclusion without drift?
3. Is problem_finally fully self-contained?
4. Is the resulting task suitable for Lean formalization?

Review scope.
You must check:
- whether any assumption has been lost;
- whether any assumption has been changed;
- whether any quantifier or logical direction has changed;
- whether any domain restriction has changed;
- whether any conclusion has been dropped;
- whether any conclusion has been added;
- whether any required dependency is missing;
- whether any technical term or function is used without definition;
- whether the task is actually formalizable in Lean.

Detailed review criteria.

1. Assumption preservation.
Check whether problem_finally is missing any assumption that is explicit or implicit in problem. Examples include nonzero assumptions, nonnegative assumptions, positivity assumptions, positive-definite assumptions, domain restrictions, index range restrictions, quantifier scope restrictions, existence conditions, uniqueness conditions, continuity, differentiability, integrability, measurability, boundedness, invertibility, and other regularity or well-posedness conditions. If a necessary assumption from problem is absent in problem_finally, flag missing_assumption.

2. Task preservation.
Check whether the mathematical task in problem_finally is still the same as in problem. Pay special attention to non-equivalent drift, including changing exists to forall, changing equivalence to one-way implication, dropping one direction of a two-sided result, changing the domain or scope of variables, weakening or strengthening the conclusion, removing a conclusion, or adding a new conclusion. Equivalent reformulations are acceptable only when they are genuinely mathematically equivalent. If the final version changes the task in a non-equivalent way, flag task_drift.

3. Self-containedness.
Check whether problem_finally can be understood and formalized on its own. Flag missing_dependency if the statement relies on something not actually included or specified, such as a missing tag or unresolved reference, a missing theorem, lemma, proposition, corollary, or algorithm, a missing dependency on another exercise or previous result, or a required cited fact that is not stated clearly enough to use. When judging missing_dependency, read the full text carefully and confirm that the dependency is genuinely absent, rather than merely named while its relevant mathematical content has already been restated in the problem itself. Also flag missing_dependency if problem_finally refers to a tag, Theorem X.Y, Alg. X.Y, Algorithm X.Y, part (b), or another exercise-internal reference, but does not explain or restate the referenced content clearly enough for the statement to stand on its own.

4. Missing definitions.
Check whether problem_finally uses a specialized, uncommon, or technical term without defining it. Also flag this issue if a function, operator, mapping, object, or construction is used without a definition precise enough for formalization. Examples include uncommon terms or symbols such as ill-conditioned, L(x,y), or similarly specialized notation introduced without any definition. If this happens, flag missing_def.

5. Formalizability.
Judge whether the task is suitable for Lean formalization. Usually suitable: proof-oriented mathematical statements, precisely stated claims with explicit mathematical objects and conclusions, and mathematically precise computational statements. Usually not suitable: give a description, formulate, compute, plot, sketch, discuss informally, open-ended explanatory tasks, or tasks whose result is mainly visual, rhetorical, or pedagogical rather than formally mathematical. If problem_finally contains verbs such as formulate, compute, or similar wording that is not suitable for Lean formalization as a precise theorem-style target, judge the task as not suitable for Lean formalization and choose hold. If the task itself is not suitable for Lean formalization, the overall status should be hold.
\end{promptbox1}

\subsection{Constrained Repair and Re-review Prompt}
\label{app:repair_and_re-review_prompt}

This prompt is used when the source-level reviewer classifies an instance as
repairable. The repair step is constrained to modify only
\texttt{problem\_finally}; earlier fields remain fixed as audit anchors.

\paragraph{Prompt template.}
\begin{promptbox1}
Your task is to rewrite only problem_finally, using the review result and the provided fix suggestions, so that the final statement is mathematically faithful to the original task, self-contained, and easier to formalize in Lean.

You must read the input very carefully. You will be audited after your rewrite.
Be precise, conservative, and faithful to the given instructions.

Rewrite problem_finally according to the repair suggestions.

Your goal is to produce a corrected problem_finally that:
1. preserves the original mathematical meaning of problem;
2. follows the correction instructions in value_true_issues;
3. remains consistent with problem_clean and problem_standardized_math;
4. is self-contained, mathematically precise, and suitable for Lean formalization.

You must rewrite problem_finally.
You must not change any other field.

Do not output overall_status, reason, or value_true_issues.
Do not introduce any new correction that is not required by the provided review reasons and suggestions.
Do not add extra assumptions, conclusions, definitions, or reformulations beyond what is needed to carry out the stated repair.
Do not improve style for its own sake.

You must read reason and value_true_issues carefully.
Use them as the direct authority for what needs to be fixed.

Add back the missing assumption exactly as needed, without changing the mathematical task.
Rewrite problem_finally so that it matches the original task in problem again.
Restore the correct variables, domain, quantifiers, logical direction, and conclusion.
Add the missing definition or make the relevant notion explicit, but only to the extent required by the repair suggestion.
Apply all repair suggestions together, but keep the final statement minimal and faithful.

Priority of authority:
1. problem is the source of truth for the original mathematical task.
2. value_true_issues and reason tell you exactly what must be repaired.
3. problem_standardized_math helps preserve the already standardized mathematical structure.
4. problem_clean helps remove irrelevant textbook narrative.
5. problem_with_context may be consulted only when needed to resolve ambiguity, not to invent new content.

If these sources conflict, preserve the original mathematical intent of problem and follow the explicit repair instructions.

Do not:
- modify problem, problem_clean, or problem_standardized_math;
- add new assumptions not justified by the repair feedback;
- weaken the conclusion;
- strengthen the conclusion;
- introduce a new proof strategy;
- insert hints or remarks;
- output analysis, commentary, or explanation outside the JSON.

All fields except problem_finally must remain unchanged from the input.

Before you output, verify all of the following:
1. Only problem_finally has been changed.
2. The rewrite follows the provided repair suggestions.
3. The rewritten problem_finally does not introduce any extra content beyond the required repair.
4. The rewritten problem_finally is self-contained and suitable for Lean formalization.
5. The output is valid JSON.
6. Mandatory boundary-case safeguard: explicitly check edge conditions such as nonemptiness and zero-valued cases for the repaired statement. If an edge case fails, produce a concrete counterexample in your verification and treat the item as requiring revise with a clear reason in downstream review.

Your job is not to generate a better problem in general.
Your job is to repair problem_finally, and only problem_finally, so that it correctly reflects the original task and the given review suggestions.
\end{promptbox1}

\subsection{Compact Translator Prompt}
\label{app:compact-translation-prompt}

This prompt is used for the initial JSON-to-Lean translation stage. It translates
one semantic block at a time into a Lean declaration while reusing previously
accepted declarations from the frozen context.

\paragraph{Prompt template.}
\begin{promptbox1}
You are a JSON-to-Lean block translator.

Input.
- current JSON semantic block with fields such as index, source, source_idx, kind, term, content, and problem;
- section_context: previously accepted Lean declarations in the same namespace;
- optional mcp_context or external source context.

Task.
Translate exactly one JSON semantic block into one Lean 4 declaration, plus helper declarations only when necessary. Reuse existing declarations from section_context whenever possible. Output only the Lean code fragment for the current block.

Use semantic content in this priority order:
1. content;
2. problem;
3. term, only as a naming or contextual hint.

Declaration rules by kind:
- defn: produce one source-faithful Lean def. Do not use sorry or vacuous bodies.
- thm: produce exactly one theorem as the primary declaration. Keep all source assumptions explicit. Use a concise, semantic, Lean-safe name. The proof body must be exactly "by sorry".
- algo: produce one structure encoding source-grounded algorithmic state, parameters, and invariants. Do not encode theorem conclusions as fields.
- opt_prob: produce one structure encoding decision variables, objectives, constraints, and source-grounded optimization components.

Context reuse rules.
Before generating code, scan section_context. Reuse matching definitions, structures, theorem names, notation, helper predicates, and directives. Do not redefine equivalent symbols unless the meaning differs.

Lean and safety constraints:
- Output only Lean code for insertion inside the shared namespace.
- Do not output imports, namespace wrappers, closing "end"s, block comments, Markdown fences, JSON/YAML, or explanations.
- Do not introduce vacuous placeholders such as "def X := True", empty Prop wrappers, or meaningless predicates.
- Do not use axiom, admit, unsafe, opaque, or fake mathematical content.
- For theorem blocks, the only allowed unfinished proof is "by sorry".
- Preserve source assumptions, equalities, relations, constructions, and boundary conditions.
- Do not weaken or strengthen the source statement except minimally to obtain a well-typed Lean formalization.
- Add open, open scoped, variable, or set_option directives only when needed and not already provided by section_context.
- Prefer Mathlib-native notation and fully qualified names when they avoid unnecessary directives.
- Use "fun" instead of lambda symbols.
- In structures, put each field on its own line.
- For Euclidean inner products, use the notation supported by the current Lean environment; do not add unsupported scalar-field subscripts.

Final check.
The output must preserve the source intent, reuse available context, match the required declaration shape for the block kind, avoid placeholders, and contain only the Lean fragment for the current block.
\end{promptbox1}

\subsection{Pass@$k$ Proof-Generation Prompts}
\label{app:passk-prompts}

These prompts are used in the non-agentic pass@$k$ experiments. Each model
receives a Lean block with one designated proof placeholder,
\texttt{FILL\_PROOF\_HERE}, and returns only the proof body that replaces it.
Other \texttt{sorry} tokens in the template are left untouched.

\subsubsection{System Prompt}

\paragraph{Prompt template.}
\begin{promptbox1}
You are a Lean 4 proof assistant. Your task is to fill in exactly one proof placeholder.

Rules.
- Output only the Lean proof term or tactic body that should replace the /- FILL_PROOF_HERE -/ marker.
- Do not output the surrounding theorem, lemma, or def block.
- Do not output thinking, reasoning, chain-of-thought, analysis, or any hidden deliberation; output the code directly.
- Do not output imports, namespaces, sections, declarations, comments, explanations, or Markdown outside one Lean code block.
- The code you output will be inserted verbatim at the marker. It must fit exactly at that position.
- Do not modify, repeat, comment out, delete, rename, weaken, or otherwise alter the original theorem, lemma, or def statement, hypotheses, target, imports, namespace, or surrounding code.
- Do not touch any other sorry tokens already present in the template; they belong to later steps.
- Do not add or use axiom, admit, unsafe, opaque, or proof-search/suggestion commands such as apply?, exact?, rw?, simp?, aesop?, solve_by_elim?, or library_search.
- Never use sorry.
- Before using any lemma, theorem, or definition name, verify it is resolvable in the current context.
- Never guess or invent lemma names based on similarity.
- If a needed lemma is not resolvable from the search results, either switch to available lemmas or define a helper first without sorry, axiom, admit, unsafe, or opaque.
- Wrap your code in a Lean code block.
\end{promptbox1}

\subsubsection{First-Turn User Prompt}

\paragraph{Prompt template.}
\begin{promptbox1}
Here is a Lean 4 block containing a proof placeholder /- FILL_PROOF_HERE -/ and possibly other sorry tokens, which you must leave untouched.

Lean block.

{block_code}

Available tool/search context.

{mcp_sections}

Output only the Lean proof term or tactic body that should replace /- FILL_PROOF_HERE -/.

Do not output thinking, reasoning, chain-of-thought, analysis, or any hidden deliberation; output the code directly.
Do not output the theorem, lemma, or def block itself.
The proof body will be inserted verbatim at the marker.

If the required lemma or definition is not found in the search results, define it yourself first before using it.
Use exact identifier names from searchable context; do not use similar-looking names that are not resolvable.

Do not use sorry, axiom, admit, unsafe, opaque, apply?, exact?, rw?, simp?, aesop?, solve_by_elim?, or library_search.
Return only the proof body in a Lean code block.
\end{promptbox1}

\subsubsection{Retry-Turn User Prompt}

\paragraph{Prompt template.}
\begin{promptbox1}
Your previous proof body produced compile errors.
Here is the original template, your last proof body, and the compiler diagnostics.

Original template with placeholder.

{block_template}

Your previous proof body.

{prev_code}

Compile errors.

{compile_errors}

Available tool/search context.

{mcp_sections}

Output only the corrected Lean proof term or tactic body that should replace /- FILL_PROOF_HERE -/ in the original template.

Do not output thinking, reasoning, chain-of-thought, analysis, or any hidden deliberation; output the code directly.
Do not output the theorem, lemma, or def block itself.
The proof body will be inserted verbatim at the marker.
Keep any other sorry tokens unchanged by not mentioning them.

If the required lemma or definition is not found in the search results, define it yourself first before using it.
Before writing the final proof, ensure every new identifier you use is resolvable; otherwise revise the approach.

Do not use sorry, axiom, admit, unsafe, opaque, apply?, exact?, rw?, simp?, aesop?, solve_by_elim?, or library_search.
Do not add comments, explanations, or text outside the Lean code block.
Output only the corrected proof body in a Lean code block.
\end{promptbox1}

\section{Limitations}
CAM-Bench has several limitations. First, although CAM-Bench targets computational and applied mathematics, its source material is still optimization-centric, with coverage mainly in optimization, numerical linear algebra, and numerical analysis. Other areas of applied mathematics, such as PDEs, control, and scientific computing, are not yet systematically covered. 

Second, benchmark construction uses an LLM-assisted pipeline for dependency recovery, normalization, Lean formalization, and semantic review. We reduce this risk through source-grounded dependency expansion, constrained repair, Lean compilation, and semantic validation. However, subtle mismatches, such as boundary cases may still remain.

Finally, due to computational cost, the external agentic systems are evaluated on 200 uniformly sampled instances rather than the full benchmark. These results are informative but should be interpreted as sampled estimates rather than complete measurements over all benchmark instances.

\section{Boarder Impacts}

CAM-Bench aims to support reliable evaluation of mathematical reasoning systems by providing Lean-verified proof targets in computational and applied mathematics. By focusing on context-dependent textbook problems, it helps diagnose failures in assumption tracking, library grounding, type discipline, and long-horizon proof decomposition. The benchmark may encourage progress in formal applied mathematics, including stronger Mathlib infrastructure, more capable theorem-proving agents, and more faithful autoformalization pipelines. It also complements existing formal benchmarks that primarily emphasize Olympiad-style problems, algebraic domains, or isolated theorem statements. CAM-Bench should therefore be used as one component of a broader evaluation suite. Strong performance on this benchmark indicates progress in Lean-based proof generation for context-dependent applied mathematics.

\end{document}